\documentclass[10pt,twocolumn,letterpaper]{article}
\usepackage[pagenumbers]{cvpr}

\usepackage{amsmath,amsfonts,bm}

\def\eqref#1{equation~\ref{#1}}

\def\1{\bm{1}}

\DeclareMathAlphabet{\mathsfit}{\encodingdefault}{\sfdefault}{m}{sl}
\SetMathAlphabet{\mathsfit}{bold}{\encodingdefault}{\sfdefault}{bx}{n}

\definecolor{cvprblue}{rgb}{0.21,0.49,0.74}
\usepackage[pagebackref,breaklinks,colorlinks,allcolors=cvprblue]{hyperref}
\usepackage[capitalize]{cleveref}
\usepackage{url}
\usepackage{graphicx}
\usepackage{adjustbox}
\usepackage{subcaption}

\usepackage{booktabs,tabularx}
\usepackage{array}          %
\usepackage{pifont}         %

\newcommand{\ours}{VPBench}

\newcolumntype{Y}{>{\centering\arraybackslash}X}

\usepackage{xspace}
\def\paperID{} %
\def\confName{CVPR}
\def\confYear{2026}

\title{Visually Prompted Benchmarks Are Surprisingly Fragile}
\author{
Haiwen Feng$^{*}$ \quad
Long Lian$^{*}$ \quad
Lisa Dunlap$^{*}$
\\
Jiahao Shu \quad
XuDong Wang \quad
Renhao Wang \quad
Trevor Darrell \quad
Alane Suhr \quad
Angjoo Kanazawa
\\
UC Berkeley
\\
{\footnotesize $^{*}$Equal contribution.}
}

\begin{document}

\twocolumn[{
    \maketitle
    \vspace{-0.5cm}
    \begin{center}
        \centering
        \includegraphics[width=\linewidth]{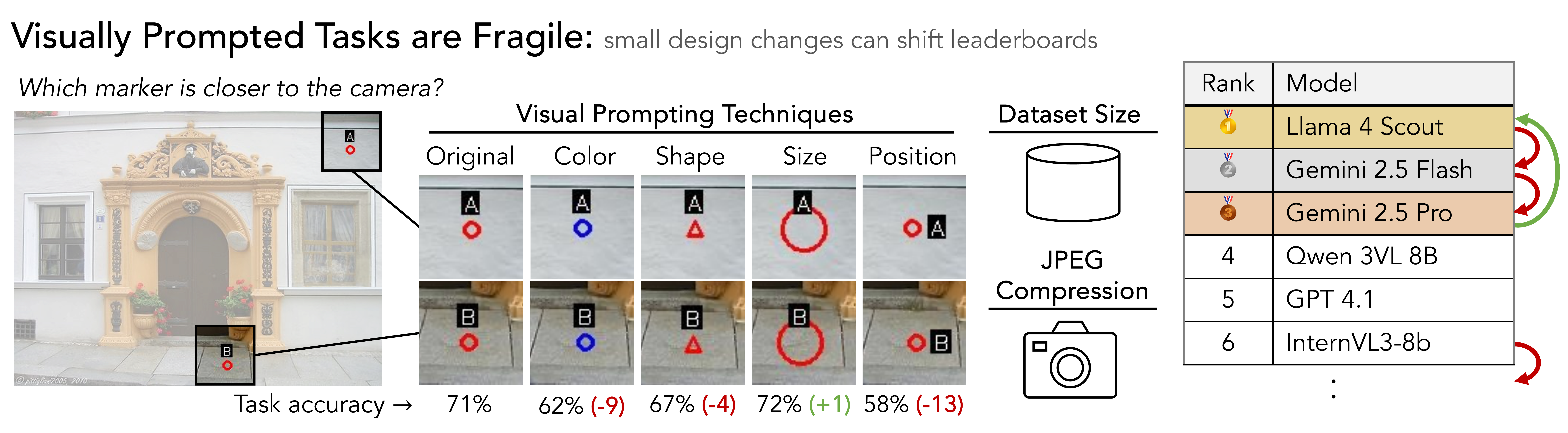}
        \vspace{-0.6cm}
        \captionof{figure}{\textbf{Small, seemingly irrelevant changes in visual prompting dramatically alter VLM predictions.}
        \textbf{Left:} Qwen2.5-VL accuracy under different visual marker variants on BLINK relative depth. Changes in marker size, shape, color, and label position lead to significant accuracy swings up to 13\%. 
        \textbf{Right:} Leaderboard of VPBench Relative Depth. Such variations can reorder leaderboards, with model rankings shifting even when nothing about the underlying task changes.}
        \label{fig:teaser}
    \end{center}
}]

\begin{abstract}

A key challenge in evaluating VLMs is testing models' ability to analyze visual content independently from their textual priors. Recent benchmarks such as BLINK probe visual perception through \textit{visual prompting}, where questions about visual content are paired with coordinates to which the question refers, with the coordinates explicitly marked in the image itself.  While these benchmarks are an important part of VLM evaluation, we find that existing models are surprisingly fragile to seemingly irrelevant details of visual prompting: simply changing a visual marker from red to blue can completely change rankings among models on a leaderboard. By evaluating nine commonly-used open- and closed-source VLMs on two visually prompted tasks, we demonstrate how details in benchmark setup, including visual marker design and dataset size, have a significant influence on model performance and leaderboard rankings.
These effects can even be exploited to \textbf{lift weaker models above stronger ones}; for instance, slightly increasing the size of the visual marker results in open-source InternVL3-8B ranking alongside or better than much larger proprietary models like Gemini 2.5 Pro. We further show that low-level inference choices that are often ignored in benchmarking, such as JPEG compression levels in API calls, can also cause model lineup changes. These details have substantially larger impacts on visually prompted benchmarks than on conventional semantic VLM evaluations. To mitigate this instability, we curate existing datasets to create \ours, a larger visually prompted benchmark with 16 visual marker variants. 
We open-source \ours{} and our analysis framework at: \url{https://lisadunlap.github.io/vpbench/}.

\end{abstract}

\section{Introduction}

Despite the rapid progress of vision-language models (VLMs), their visual perception capabilities remain underexplored. Most existing benchmarks conflate visual understanding with language priors and factual recall, making it unclear whether models genuinely perceive or merely retrieve. 
To address this, visual prompting has emerged as a targeted paradigm: by marking regions in an image and posing spatial or perceptual questions, as originally proposed by \cite{Krishna2016VisualGC, Zhu2015Visual7WGQ} and made most relevant as a VLM benchmark with the BLINK dataset~\cite{fu2024blink}, these perceptual tasks assess low-level visual understanding that humans solve effortlessly, in contrast to the knowledge-centric reasoning required by benchmarks such as MME or MMMU~\cite{fu2023mme,yue2024mmmu}.

However, within this visually prompted evaluation regime, we find that model performance is surprisingly sensitive to seemingly minor design choices in the benchmark itself. As illustrated in Figure 1, variations in the size, style, or layout of visual markers can substantially affect accuracy and even reorder model rankings. Beyond prompt design, incidental implementation details, such as random sample selection, image compression settings, or floating-point precision, can further contribute to this instability. Many of these design choices are inherited from conventional, knowledge-focused VLM benchmarks, where such factors have minimal influence and are thus treated as inconsequential. Yet, in visually prompted evaluations, these non-semantic elements become hidden confounders, capable of markedly distorting model performance and leaderboard rankings. Consequently, existing visually prompted benchmarks exhibit an inherent fragility—blink again, and an apparently incidental change can shift reported scores, echoing the formatting sensitivities observed in LLMs~\citep{sclar2023quantifying}. 
Such instability undermines confidence in benchmark-driven progress and echoes recent concerns about leaderboard fragility in both language and vision domains.

We explore three such sources of evaluation instability across nine modern VLMs on BLINK~\citep{fu2024blink} as well as two more large scale datasets, \ours{}-RD and \ours{}-SC, that we curate using DA2K~\citep{depth_anything_v2}, and SPair~\citep{min2019spair}. 
First, we demonstrate the effect of \textit{sample choice}: random resampling of image subsets, matched in its modest size to BLINK and drawn from a fixed pool of visually prompted tasks, leads to substantially reordered model rankings, despite the subsets being statistically indistinguishable in size and difficulty. Second, we examine \textit{visual prompt formatting}. On four visually prompted datasets, we evaluate our models with 16 different marker styles varying in size, shape, color, and label placement. We find that marker style can cause accuracy swings of up to 21\% on the same image-question pairs, often causing ranking reversals among state-of-the-art models. Finally, we show that \textit{inference-time implementation details} which are imperceptible to humans such as JPEG compression further perturb results in statistically significant ways. Additionally, we find that this is specific to visually prompted tasks, as applying the same intervention to more traditional VLM benchmarks does not significantly change the results. We also demonstrate how this fragility can be exploited to ``game'' leaderboards. For example, strategically selecting the visual marker to be a square instead of a circle causes a weaker model like InternVL3-8B to rank above stronger models like Gemini 2.5 Pro on the BLINK relative depth estimation task. 

These findings suggest that much of the variation among model performance reported on vision-language benchmarks comes not from differentiated intrinsic capabilities of grounding language in vision, but from incidental details of prompting, sampling, and implementation. 
To address this, we release the expanded visually prompted datasets as \ours{}. \ours{} is a benchmark covering 16 different visual marker variants, and overall boosting the dataset size from 224 samples in BLINK relative depth and semantic correspondence to 35,088 annotated images across relative depth (\ours{}-RD) and semantic correspondence (\ours{}-SC). Lastly, we provide suggestions on how to more robustly evaluate  visually prompted tasks and guidance on when to trust the leaderboard rankings. We release our proposed~\ours{} along with our inference code, which supports varying visual markers and image compression settings, as a reference inference for stable evaluation at \url{https://lisadunlap.github.io/vpt-fragile/}.

\section{Related Work}
\paragraph{Perception-focused VLM benchmarks.}
Our work builds on efforts to separate low-level VLM perception from high-level reasoning. BLINK, for example, recasts classic vision tasks into visually prompted questions, showing that VLMs struggle on problems humans solve ``in a blink'' \citep{fu2024blink}, motivating follow-up work on perception-augmented representations \citep{bigverdi2025perception}. Orthogonal work evaluates robustness via controlled input variations, such as perturbations to image–text pairs, programmatic generation of task variants, geometric invariances, or spatio-temporal video manipulations \citep{park2024rococo, zou2024dynamath, fan2025unveiling, agarwal2024mvtamperbench}. These benchmarks highlight VLM gaps in both perception and robust invariance.

\paragraph{Text and \emph{visual} prompt sensitivity.}
Prompt design can dominate model performance. In language models, small, meaning-preserving formatting tweaks can significantly swing accuracy~\citep{sclar2023quantifying}. This sensitivity extends to the \emph{visual} prompts in VLMs, where the choice of marker (e.g., a dot vs. a box) can alter model attention and outcomes \citep{shtedritski2023does}. Indeed, the space of visual prompts has itself become an optimization target to boost accuracy~\citep{zhang2025autov}. These findings imply that evaluations using a single visual style may reflect prompt idiosyncrasies more than true model competence.

\paragraph{Evaluation instability: leaderboard fragility and implementation subtleties.}
Evaluation outcomes for VLMs are highly unstable, sensitive both to benchmark design and to low-level implementation details. 
\emph{On the benchmark side}, seemingly minor factors such as altering multiple-choice option order or varying random seeds can flip leaderboard rankings \citep{alzahrani2024benchmarks, madaan2024quantifying}. Community leaderboards can also be gamed via selective submissions and feedback loops~\citep{singh2025leaderboard}, motivating automated pipelines that continuously refresh and diversify test sets \citep{li2024crowdsourced}. 
\emph{On the implementation side}, minor choices can likewise skew results: in image generation, resizing filters or JPEG compression significantly impact FID scores \citep{parmar2021cleanfid}, while in multimodal evaluation, imperceptible perturbations or numerical precision differences (FP16 vs. FP32) can destabilize outputs and compromise reproducibility \citep{he2025nondeterminism, vice2025reliability, yuan2025give}. Together, these findings underscore that leaderboard orderings may reflect artifacts of evaluation pipelines as much as genuine model ability.

\paragraph{Spatial reasoning benchmarks and methods.}
Spatial reasoning has been a long-standing challenge, from early synthetic benchmarks \citep{johnson2017clevr, suhr2018corpus} to modern evaluations. Recent studies consistently show that state-of-the-art VLMs fail on simple spatial tasks, often performing near chance on benchmarks probing relative positioning and grounding \citep{kamath2023s, wang2024picture, stogiannidis2025mind}. To close this gap, a new wave of work introduces spatially-aware architectures and large-scale, spatially-grounded training data \citep{chen2024spatialvlm, cheng2024spatialrgpt, deng2025internspatial}.

\textbf{In summary}, while prior work has documented fragility in \emph{textual} prompting and robustness under broad visual corruptions, we focus on the \emph{fine-grained, visually prompted} regime—benchmarks that explicitly mark regions or points in an image to elicit low-level perceptual judgments (e.g., relative depth). We demonstrate that (i) visual prompt style is a primary confounder, (ii) i.i.d.\ resampling from a fixed superset can reorder model rankings
, and (iii) low-level implementation choices exacerbate variance. To counteract this, we advocate confidence-aware reporting across diversified visual prompts and stratified resamplings, analogous in spirit to FormatSpread’s multi-format evaluation~\cite{sclar2023quantifying}, but tailored to VLM perception, which yields markedly more stable rankings on BLINK-like tasks.

\section{Visually Prompted Tasks and Probing Them}

\begin{figure}[h]
    \centering
    \includegraphics[width=\linewidth]{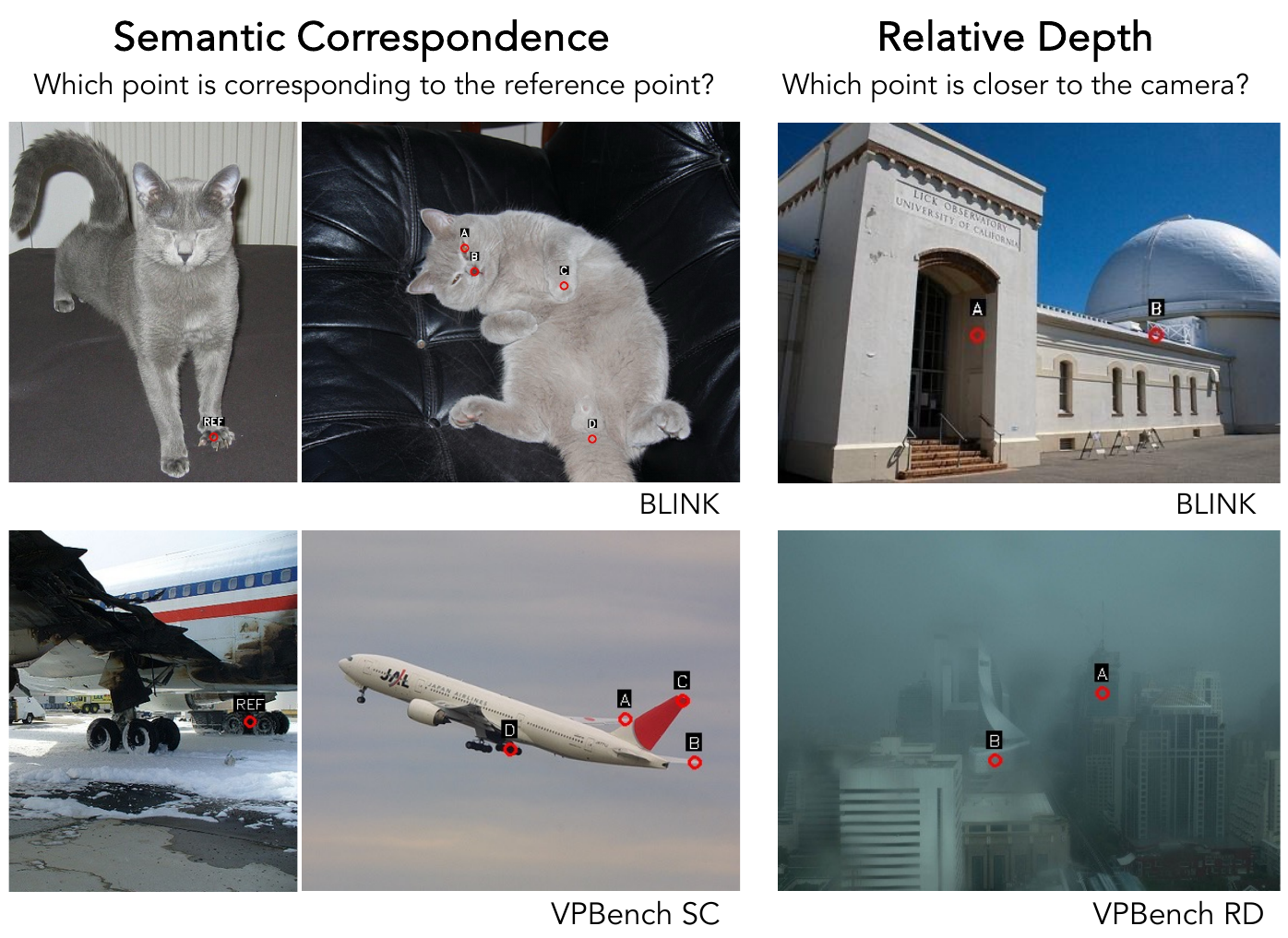}
    \caption{\textbf{Examples of visually prompted tasks.} Visually prompted tasks (VPTs) place visual markers in the image to ask questions such as relative depth and semantic correspondence.}
    \vspace{-1em}
    \label{fig:tasks}
\end{figure}

\subsection{What are Visually Prompted Tasks?}

A \textit{visually prompted task} explicitly marks regions of an image and asks about relationships within or between images~\cite{fu2024blink}. %
We pick two typical visually prompted tasks (Figure~\ref{fig:tasks}). In \textit{semantic correspondence},  two images with marked points are shown and the model is asked which marker corresponds to the same object part or region as the reference marker.
In \textit{relative depth}, two locations in an image are marked with ``A'' and ``B,'' and the model is asked which is closer to the camera.

Functionally, such visual prompting complements verbal prompting, serves as an intuitive and effective interface for querying or referencing fine-grained visual content. Humans naturally point to a location on a map, for example, rather than verbally describing its longitude and latitude. Meanwhile, unlike most VLM tasks that depend on broad world knowledge (e.g., MMMU), these visually prompted tasks are perceptual: humans solve them ``in the blink of an eye,'' relying on fine-grained visual reasoning rather than factual recall. This makes them a natural choice for evaluating how well models recognize objects and spatial relationships without external knowledge. BLINK~\cite{fu2024blink}, the benchmark we study, is built around such tasks and has become a de-facto standard for measuring visual perception in VLMs.
\subsection{Experimental Setup}
The BLINK dataset~\cite{fu2024blink} systematizes these visually prompted tasks via curated image--question pairs with region-level markings. It contains relative depth and correspondence tasks with high annotation quality. However, we notice that within BLINK, visual markers are not standardized: some examples use dots, others boxes or arrows. When these seemingly cosmetic differences are altered, we notice non-trivial change in model performance. This motivated our investigation into how VLM benchmarks may be fragile given seemingly irrelevant design choices.

As will be detailed in Section~\ref{sec:size}, BLINK’s small size, approximately 100 examples per split, makes it difficult to separate true performance variation from sampling noise. To reduce this ambiguity and enable more robust analysis, we curate a new benchmark called \ours{}. \ours{} is sourced from
 two larger datasets that follow the same visually prompted task format: DA2K~\cite{depth_anything_v2} and SPair-71K~\cite{min2019spair}, with the former used for relative depth estimation~(\ours{}-RD) and the latter for semantic correspondence~(\ours{}-SC). DA-2K, originally introduced for depth annotation, contains thousands of images with dense geometric labels; we repurpose it into a relative-depth task by converting pixel-level depth into pairwise comparisons following BLINK’s protocol. SPair-71K (SPair), designed for semantic correspondence, provides tens of thousands of image pairs with detailed keypoint matches; by adopting BLINK’s prompting style, we frame SPair as a visually prompted correspondence benchmark that bridges traditional CV evaluation with modern VLM usage. We then apply all interventions for the following sections to all 4 datasets. 

Additionally, we compare the instability of rank across different dataset sizes and JPEG compressions to a non visually prompted task, MME~\cite{fu2023mme}, to investigate how these interventions affect visually prompted tasks specifically. 

\begin{figure*}[htbp]
    \centering
    \begin{minipage}[b]{0.32\linewidth}
        \centering
        \includegraphics[width=\linewidth,trim={0cm 0cm 0cm 1cm},clip]{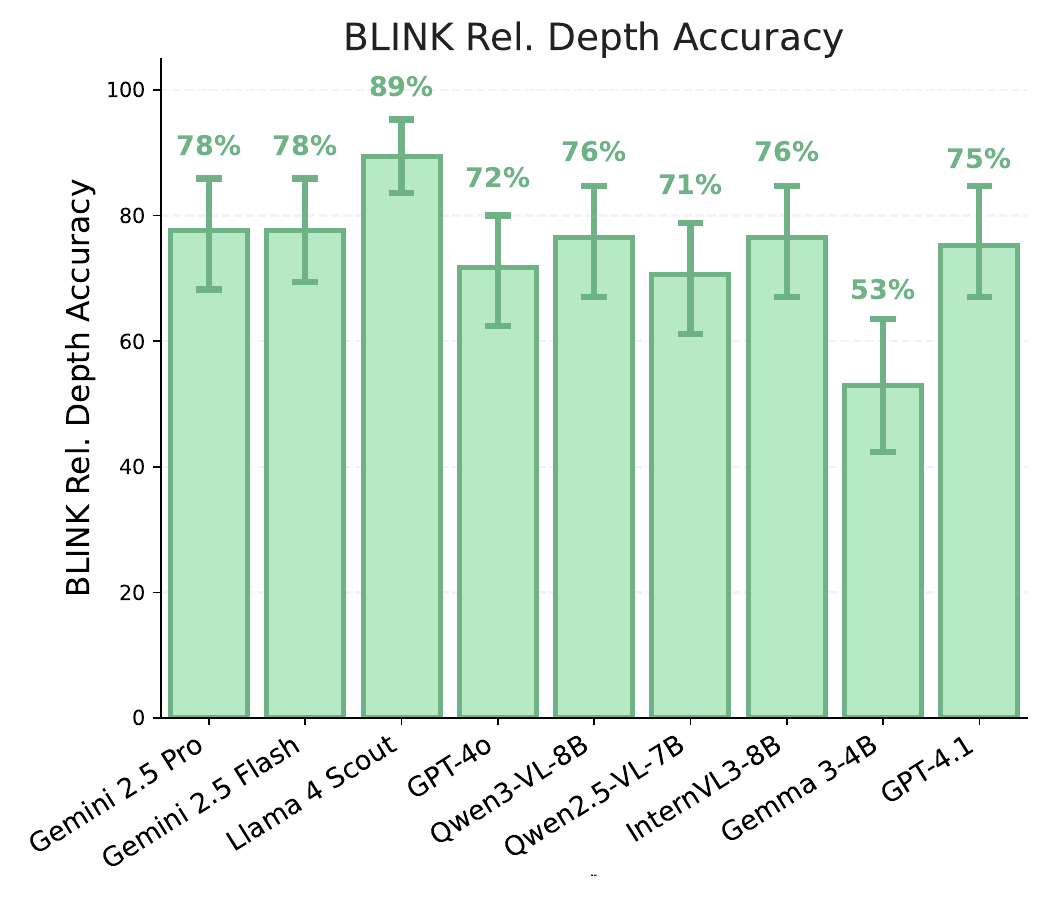}
    \end{minipage} \hfill
    \begin{minipage}[b]{0.16\linewidth}
        \centering
        \raisebox{0.65cm}{
            \includegraphics[width=\linewidth,trim={0cm 1cm 0cm 1.5cm},clip]{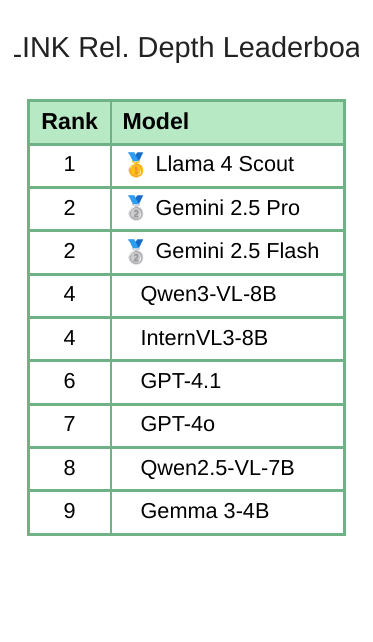}
        }
    \end{minipage} \hfill
    \begin{minipage}[b]{0.32\linewidth}
        \centering
        \includegraphics[width=\linewidth,trim={0cm 0cm 0cm 1cm},clip]{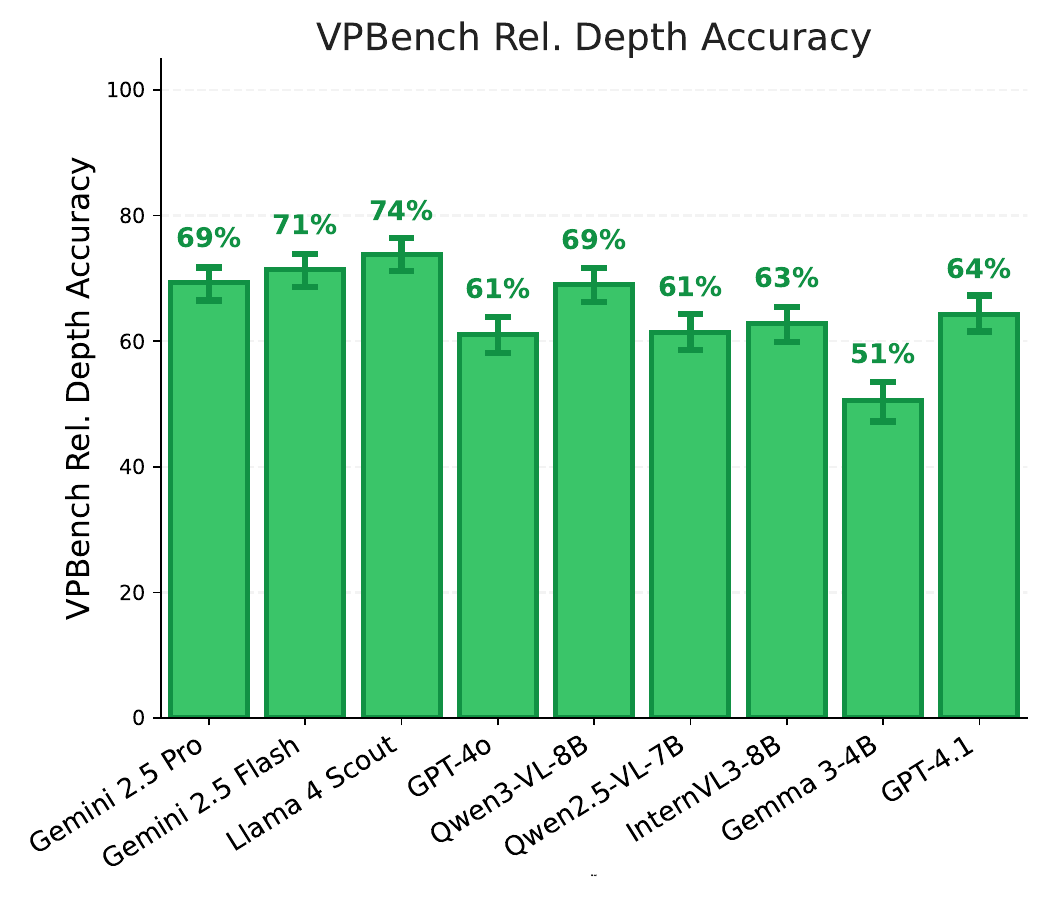}
    \end{minipage} \hfill
    \begin{minipage}[b]{0.16\linewidth}
        \centering
        \raisebox{0.65cm}{
            \includegraphics[width=\linewidth,trim={0cm 1cm 0cm 1.5cm},clip]{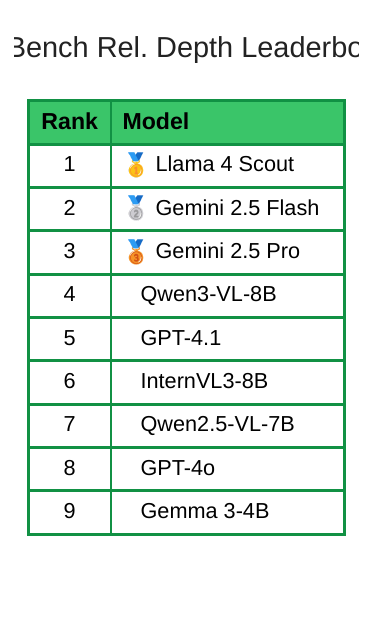}
        }
    \end{minipage}
    
    \vspace{-1em}
    
    \begin{minipage}[b]{0.32\linewidth}
        \centering
        \includegraphics[width=\linewidth,trim={0cm 0cm 0cm 2cm},clip]{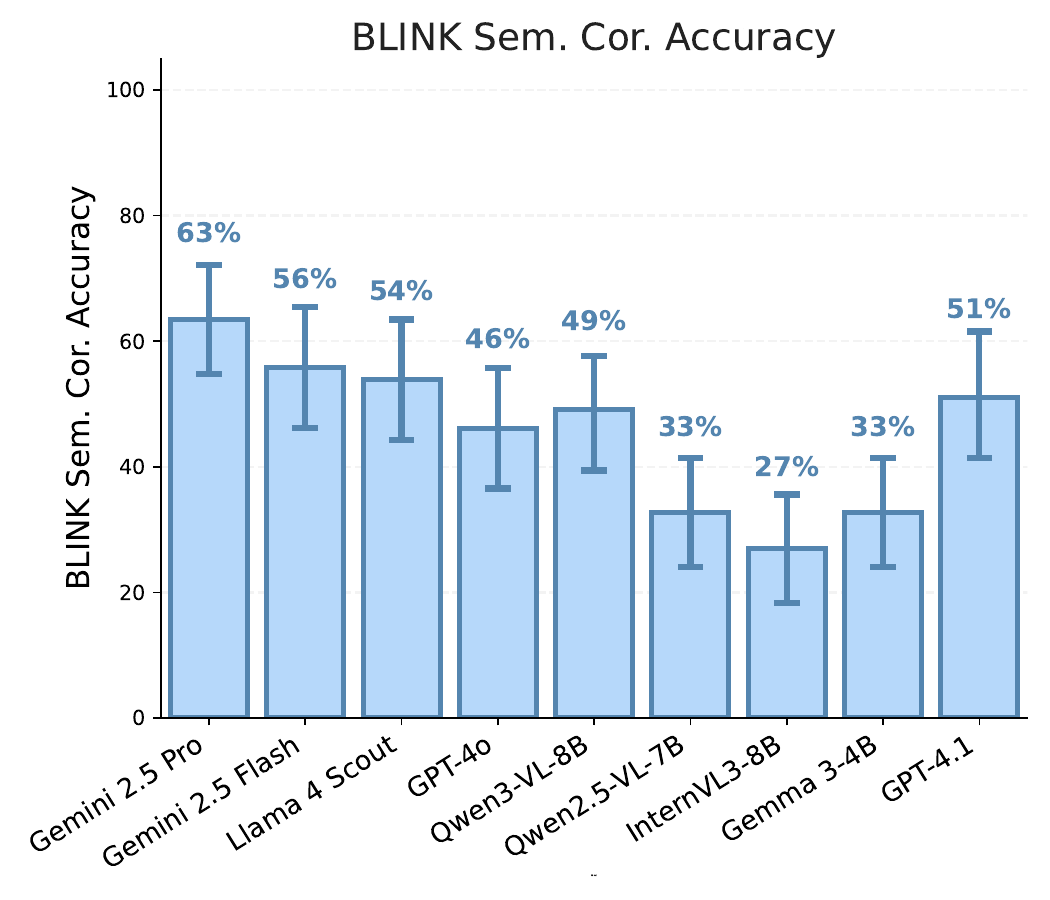}
    \end{minipage} \hfill
    \begin{minipage}[b]{0.16\linewidth}
        \centering
        \raisebox{0.65cm}{
            \includegraphics[width=\linewidth,trim={0cm 1cm 0cm 1.5cm},clip]{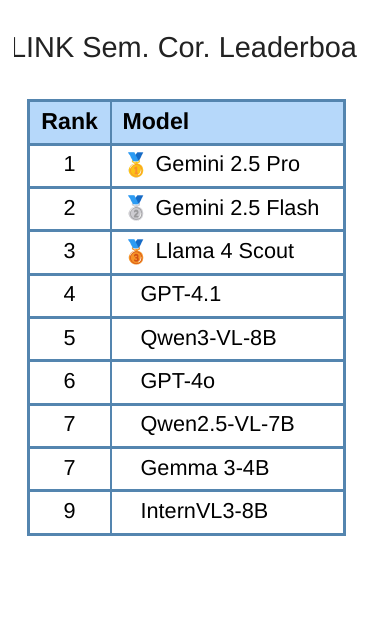}
        }
    \end{minipage} \hfill
    \begin{minipage}[b]{0.32\linewidth}
        \centering
        \includegraphics[width=\linewidth,trim={0cm 0cm 0cm 2cm},clip]{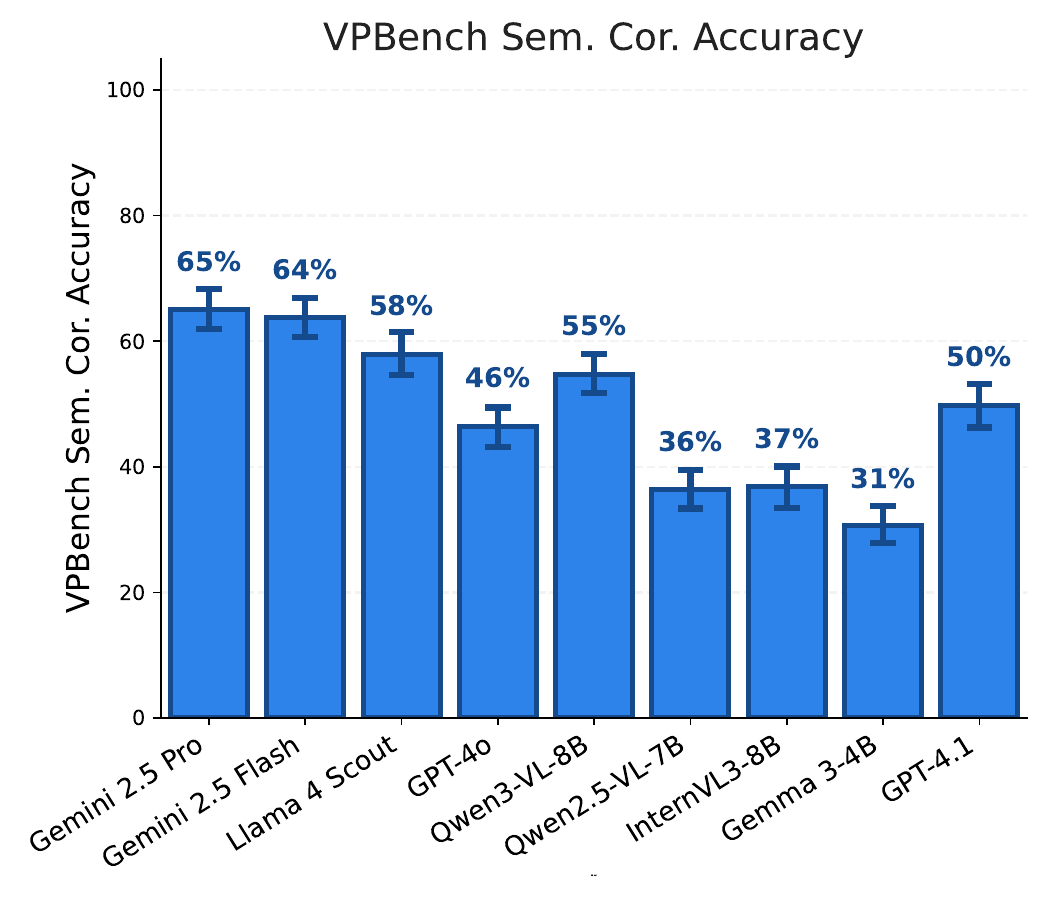}
    \end{minipage} \hfill
    \begin{minipage}[b]{0.16\linewidth}
        \centering
        \raisebox{0.65cm}{
            \includegraphics[width=\linewidth,trim={0cm 1cm 0cm 1.5cm},clip]{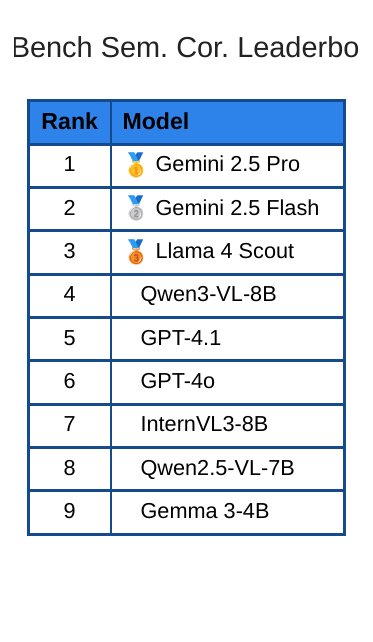}
        }
    \end{minipage}
    \caption{\textbf{Larger benchmark datasets stabilize rankings.}
Accuracies and rankings of 9 VLMs on BLINK Relative Depth, BLINK Semantic Correspondence, VPBench Relative Depth (\ours{}-RD), and VPBench Semantic Correspondence (\ours{}-SC) using BLINK's default marker convention. Error bars show 95\% confidence intervals. Compared to BLINK’s small test splits, the larger VPBench relative depth and semantic correspondence  evaluations yield substantially narrower intervals, making ranking differences easier to interpret and less sensitive to sampling noise.}
\vspace{-1em}
    \label{fig:default_acc}
\end{figure*}

We evaluate four closed-source and five open-source VLMs: Gemini 2.5 Pro, Gemini 2.5 Flash, GPT-4.1, GPT-4o, Llama 4 Scout, Qwen3-VL-8B, Qwen2.5-VL-7B, Gemma 3-4B, and InternVL3-8B. Aggregate accuracies for each benchmark appear in \cref{fig:default_acc}.

In the following section, we demonstrate the fragility of these visually prompted tasks by showing how small changes in data sampling, visual marker, and low-level implementation details can completely change the results. %

\section{Statistically Equivalent Sampling Yields Different Results}
\label{sec:size}

\begin{figure}
    \centering
    \begin{minipage}[b]{0.95\linewidth}
        \centering
        \includegraphics[width=\linewidth,trim={0cm 0cm 0cm 1cm},clip]{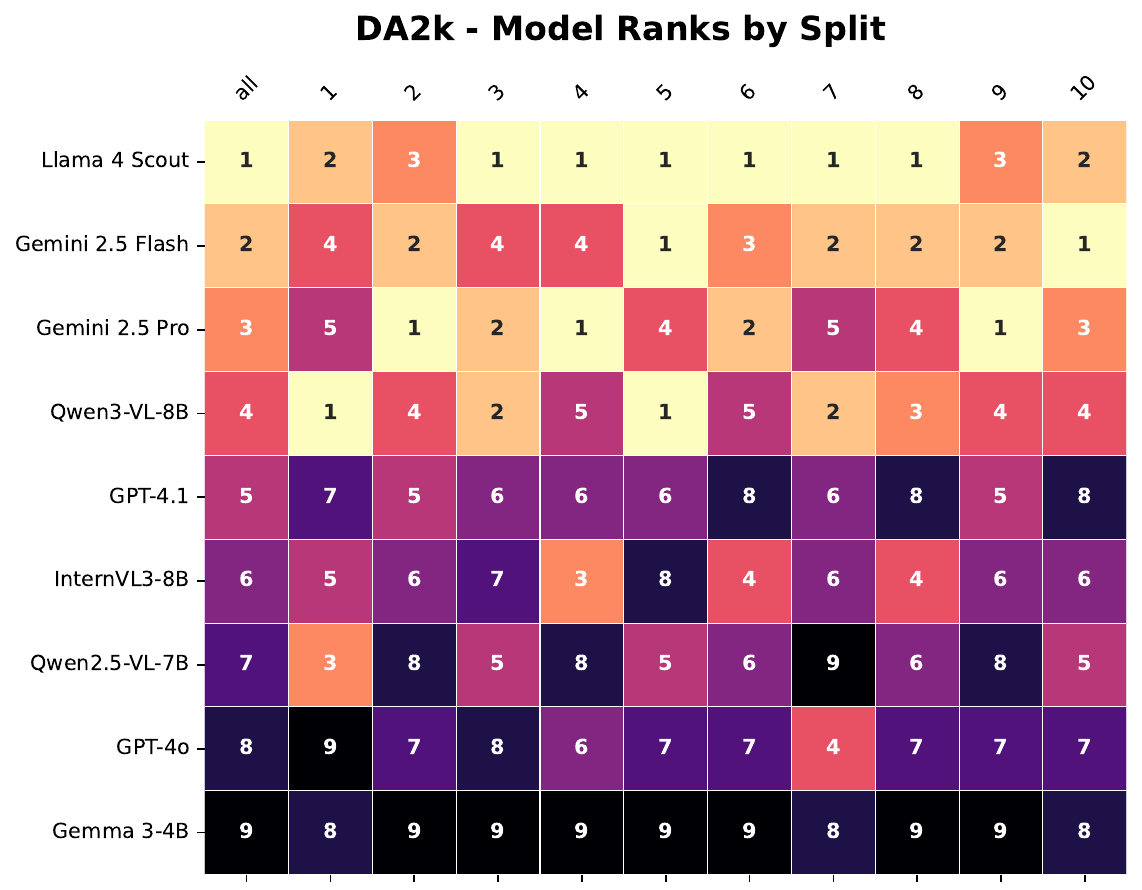}
        \subcaption{\ours{}-RD Model Ranking per data split}
    \end{minipage}
    \hfill
    \begin{minipage}[b]{0.95\linewidth}
        \centering
        \includegraphics[width=\linewidth,trim={0cm 0cm 0cm 1cm},clip]{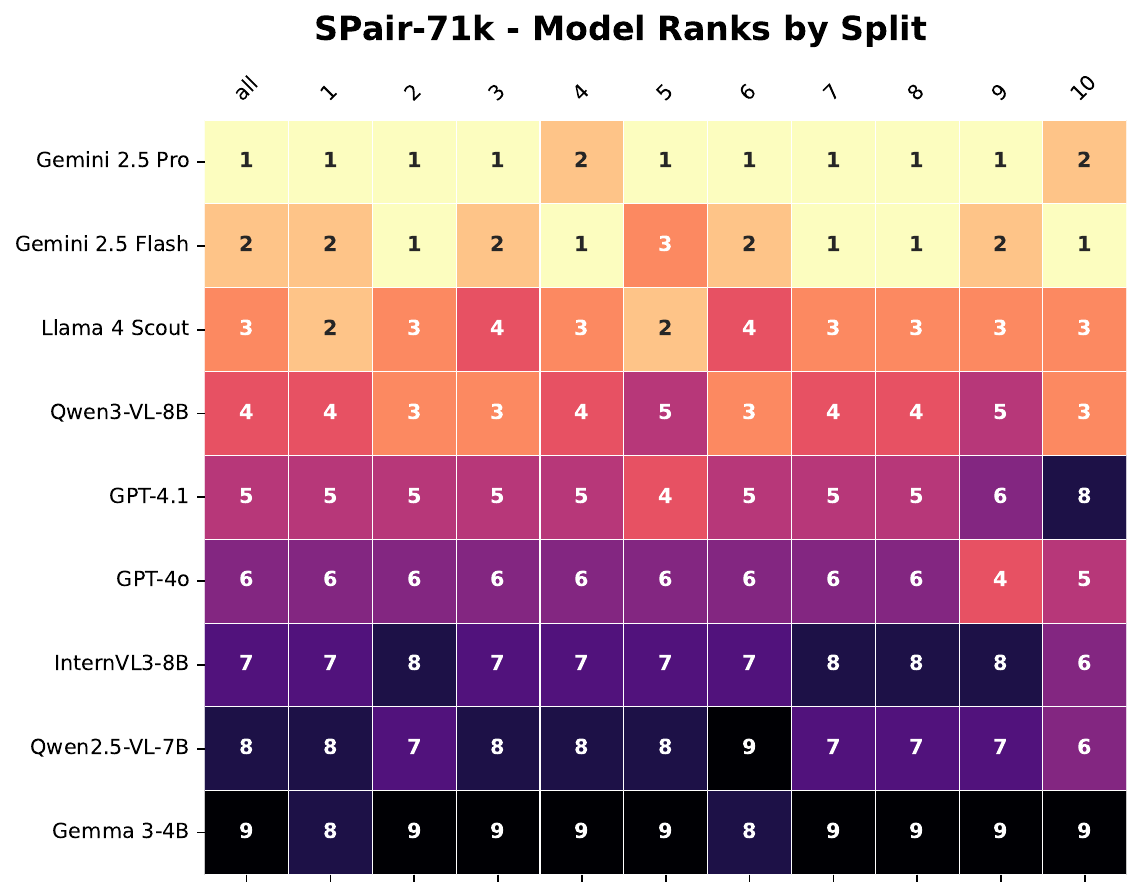}
        \subcaption{\ours{}-SC Model Ranking per data split }
    \end{minipage}
    \caption{\textbf{Model rankings across 1,000 independent 100-sample splits for \ours{}-RD and \ours{}-SC}, with the first 10 splits visualized on the x-axis, revealing substantial ranking volatility caused solely by i.i.d. resampling. 
    }
    \label{fig:data_splits}
\end{figure}

\begin{figure}[t]
    \centering
    \includegraphics[width=0.8\linewidth,trim={0cm 1cm 0cm 0cm},clip]{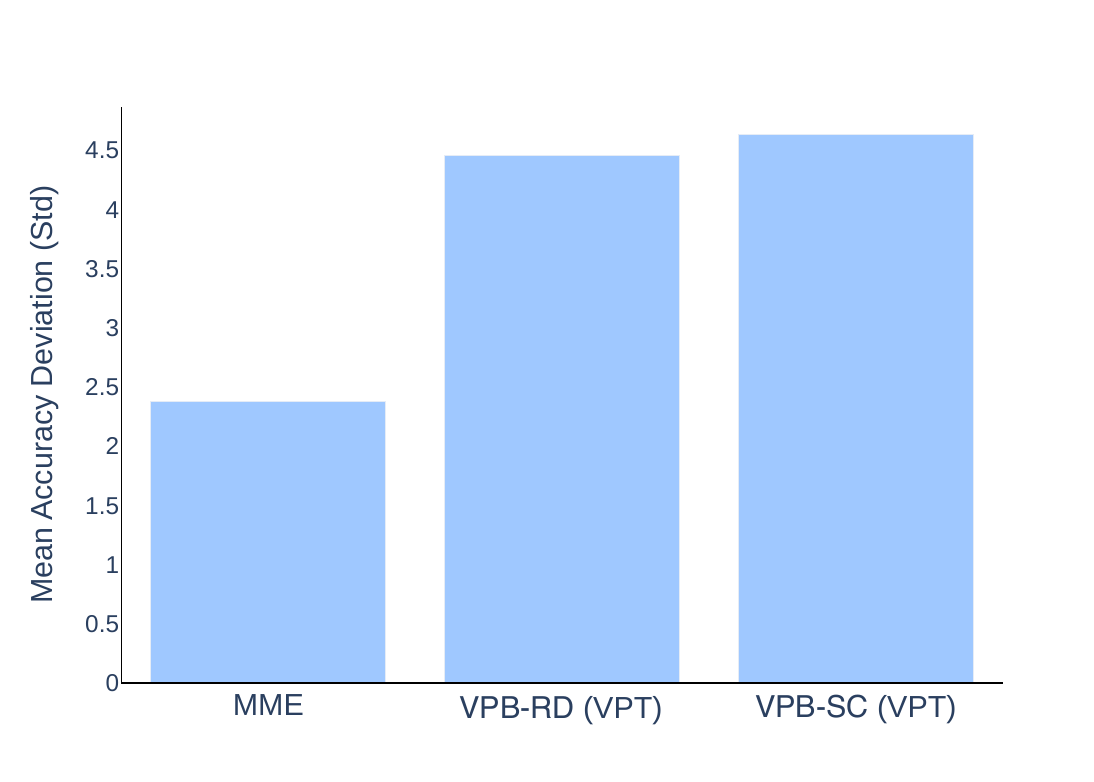}
    \caption{\textbf{Model accuracy change across 1,000 splits of 100 samples.} Standard deviation in accuracy averaged across models. We see higher rank instability across 1,000 subsets each with 100 samples for the visually prompted tasks (\ours{}-RD and \ours{}-SC) compared to knowledge-based VLM tasks (MME).}
    \label{fig:rank_correlation}
\end{figure}

When constructing a benchmark, creators typically sample a subset of data points from a large data pool, such as data collected from the Internet, to form the evaluation set. This sampling process is random, so in principle, any subset drawn from the same pool should be statistically equivalent and yield similar results. The subset size is usually fixed by convention, following prior knowledge-oriented VLM benchmarks where per-task sample counts commonly range from 50 to 500 items. While BLINK also follows this convention, we find that the random sampling step itself can meaningfully influence evaluation outcomes. If a different subset were drawn from the same underlying distribution, both the absolute accuracy and the model ranking can change noticeably.

\paragraph{Experimental setup:} We create 1,000 new BLINK size datasets by randomly sampling 100 samples from our proposed~\ours{} for both relative depth and semantic correspondence tasks, and the non-visually prompted dataset MME~\cite{fu2023mme}. Since these samples are drawn from the same underlying data distribution, the accuracies and model leaderboard should remain constant, but as shown in Figure~\ref{fig:data_splits}, we can get a complete change in rankings across \ours{}-RD and \ours{}-SC.  

\paragraph{Why does sampling matter?}
This sensitivity arises because BLINK, like many newly-released multimodal benchmarks, deliberately targets \emph{unsaturated} capabilities—i.e., tasks where performance levels are far from the ceiling, typically in the 30–60\% range rather than the 80–90\% common in relatively mature VLM tasks. Consequently, the accuracy variation in visually prompted (VP) tasks is significantly higher than in knowledge-focused benchmarks like MME (see Figure~\ref{fig:rank_correlation}). At such accuracy levels and sample sizes, even a few items can noticeably shift results; for instance, in BLINK, a 3\% accuracy change corresponds to only three additional correct responses, a margin that could easily arise by chance in a multiple-choice format. This variance is reflected in the confidence intervals in Figure~\ref{fig:default_acc}. We later highlight this impact by showing how results fluctuate as the dataset size decreases.

\paragraph{Size particularly matters for VPTs.} Additionally, we compute the mean standard deviation in model performance across splits to get numeric measures of the instability seen across splits. Figure~\ref{fig:rank_correlation} shows that the instability seen across these splits is considerably larger than that of non visually prompted tasks. This suggests that more so than for other visual tasks, visually prompted benchmarks would actually require a relatively larger number of samples to reduce the variance you see from the data. In the following sections, we will show how additional choices affect the leaderboard even when given a 10 times larger dataset size. 

\section{Visual Marker Styles Shuffle Leaderboards}
\label{sec:markers}

We investigate whether the style of the visual marker used in a prompt influences model performance, and whether it can change the relative ordering of models’ accuracy. This question is motivated by an observation of inconsistent marker styles in the BLINK benchmark dataset~\cite{fu2024blink}. The visual prompts BLINK employs as part of the question context occasionally switch in color or text positioning and we notice, for example, that simply switching a marker’s color from red to blue questions lead to measurable changes in a model’s accuracy on BLINK, suggesting current models are over-reliant on specific visual cues. In this section we create a set of marker variants and study their effects on model performance. We report results on VPBench in the main paper and report results for the BLINK subsets in the Appendix.

\begin{figure*}[t]
    \centering

    \begin{minipage}[b]{0.49\linewidth}
        \centering
        \includegraphics[width=\linewidth]{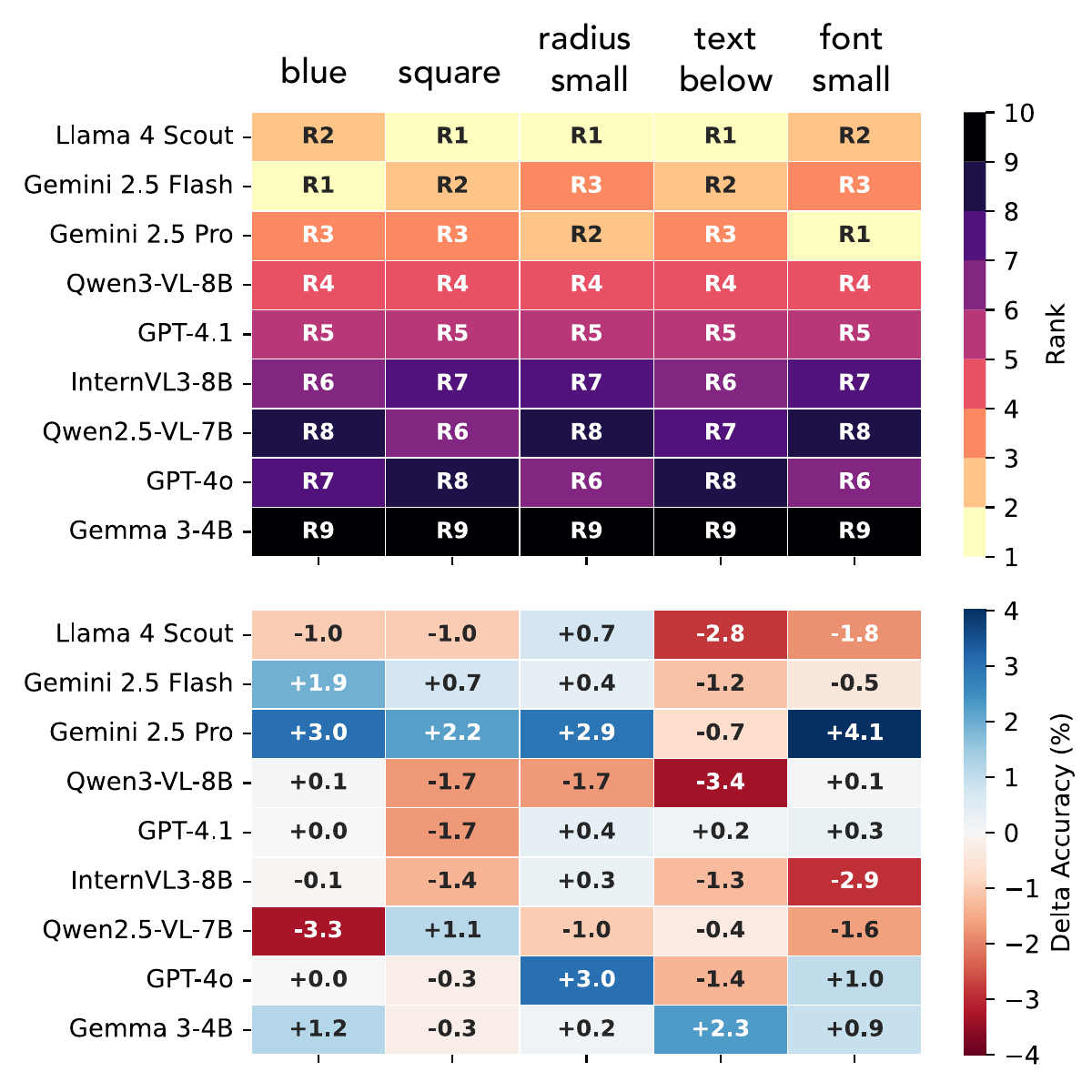}
        \subcaption{\textbf{Rank and accuracy changes across visual markers on \ours{}-RD.}}
        \label{fig:rank_marker_da2k}
    \end{minipage}
    \hfill
    \begin{minipage}[b]{0.49\linewidth}
        \centering
        \includegraphics[width=\linewidth]{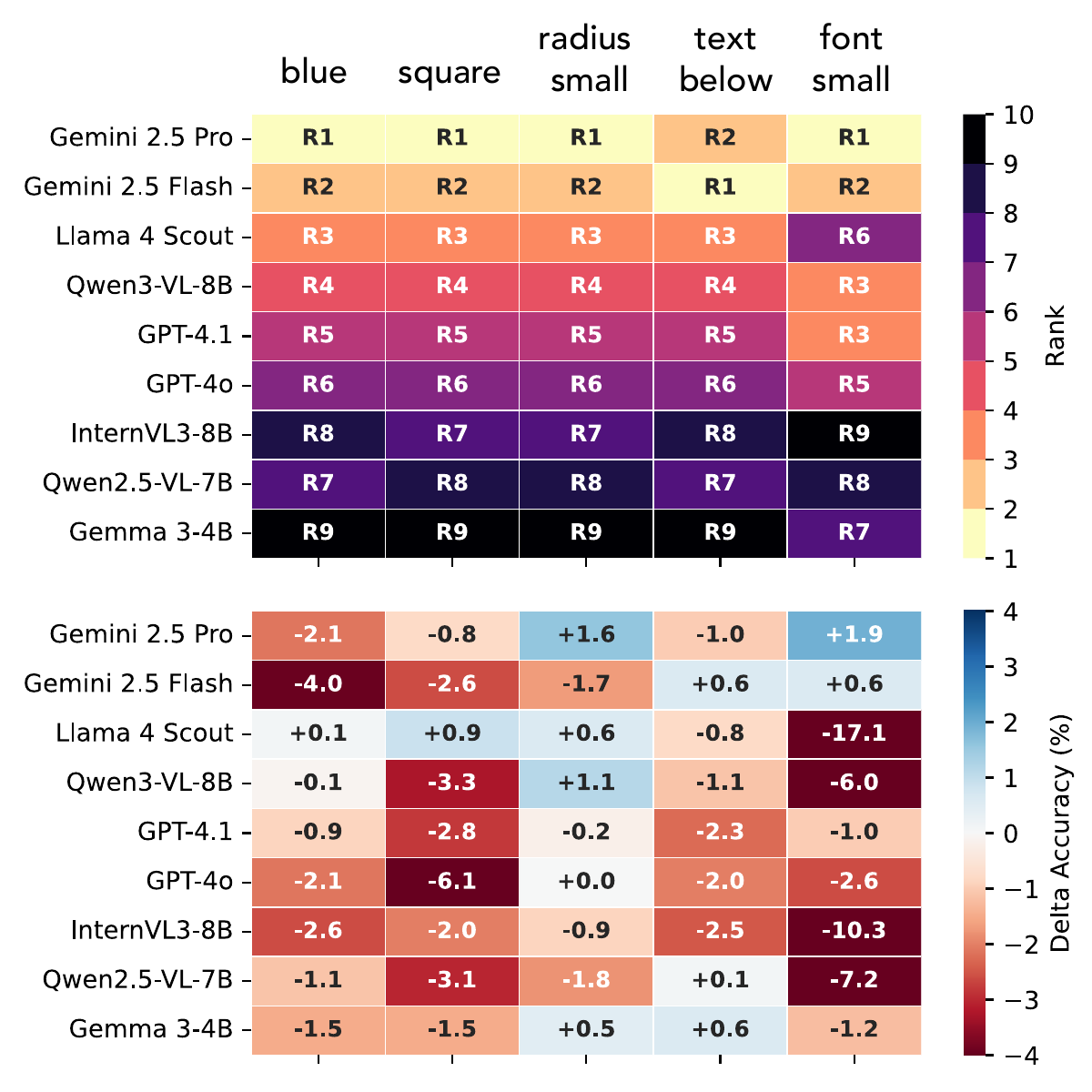}
        \subcaption{\textbf{Rank and Accuracy changes across markers on \ours{}-SC.}}
        \label{fig:rank_marker_spair}
    \end{minipage}
    \caption{\textbf{Small marker changes cause large, model-specific accuracy shifts and rank shuffles.}
Rank (top) and accuracy (bottom) variations on (a) \ours{}-RD relative depth and (b) \ours{}-SC semantic correspondence across marker variants (five representative variants shown; full 16 in Appendix). Rows are models and columns are marker styles; each cell reports the change in accuracy relative to the default BLINK marker, which uses a red circle with the label above. Positive values indicate improvements and negative values indicate degradations.}
    \vspace{-1em}
    \label{fig:marker_comparison}
\end{figure*}

\paragraph{Experimental setup:}
 Based on the most common marker seen in BLINK relative depth and semantic correspondence, we define a default marker style to be a small red circular marker with a numeric label placed above. We then create 16 alternative styles (commonly seen in visual prompting literature) spanning 6 categories: color (blue instead of red), shape (square instead of circle), marker size (larger marker, radius increased from the default to size 10), text size (larger/smaller font size), text label (1/2 VS A/B), and text position (label moved to below the marker). Figures \ref{fig:rank_marker_da2k} and \ref{fig:rank_marker_spair} display the change in accuracies and rankings of each model on relative depth and semantic correspondence tasks. 

\begin{figure}
    \centering
    \includegraphics[width=\linewidth,trim={1cm 0cm 4cm 1cm},clip]{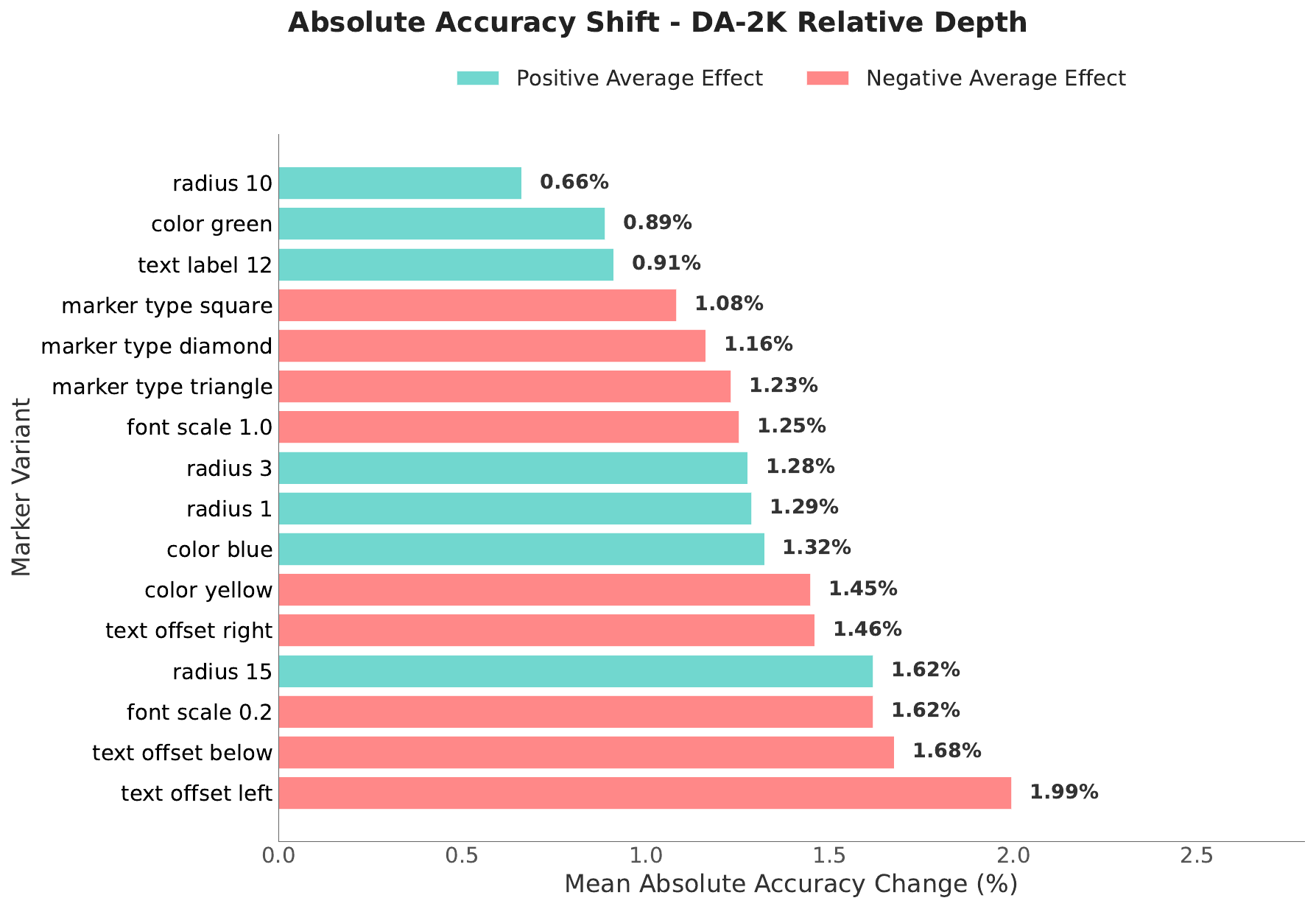}

    \caption{\textbf{Impact by marker type.} Mean absolute accuracy shift on \ours{}-RD induced by each marker variant. Changes to font scale and text placement result in the largest accuracy changes.}
    \vspace{-1.3em}
    \label{fig:maker_variance_magnitude_da2k}
\end{figure}

\paragraph{Disentangling visual marker variance from data variance.} To confirm that the variance in accuracy seen across visual markers is not simply due to the variance in the data itself, we perform a paired bootstrap to confirm that the difference between the default BLINK marker and other marker variance is statistically significant.

Let $\mathcal{D}=\{(x_\ell,y_\ell)\}_{\ell=1}^N$ be the dataset. A marker $m$ renders an annotated input $m(x_\ell)$. BLINK provides a default marker $m_0$; we also consider valid alternatives $m_1,\dots,m_K$ (e.g., dots, boxes, arrows). For a model $f$, accuracy under marker $m$ is
\[
\mathrm{acc}_{m}=\frac{1}{N}\sum_{\ell=1}^{N}\mathbf{1}\!\big[f(m(x_\ell))=y_\ell\big].
\]

To determine whether observe accuracy differences between markers are statistically significant rather than due to sampling variability, we draw $B$ bootstrap replicates $\mathcal{D}^s$ by sampling $N$ items with replacement. On each replicate, we compute accuracies with the default and an alternative marker and take the paired difference
\[
\Delta \mathrm{acc}^{\,s}_j=\mathrm{acc}^{\,s}_{m_0}-\mathrm{acc}^{\,s}_{m_j}.
\]

The distribution $\{\Delta \mathrm{acc}^{\,s}_j\}_{s=1}^B$ isolates marker variance because both conditions use the same sampled items. We test $H_0:\mathrm{acc}_{m_0}=\mathrm{acc}_{m_j}$ by checking whether zero lies in the 95\% bootstrap confidence interval of $\Delta \mathrm{acc}^{\,s}_j$.

To compare sources of variability, we report the ratio
\[
R=\frac{\mathbb{E}_j\!\left[\mathrm{Var}\big(\Delta \mathrm{acc}^{\,s}_j\big)\right]}{\mathrm{Var}\!\left(\mathrm{acc}^{\,s}_{m_0}\right)},
\]
where the denominator estimates dataset variance and the numerator estimates marker-induced variance. Values $R\!\ge\!1$ indicate that changing marker style perturbs accuracy at least as much as resampling the test set itself. We find the majority of visual markers show significant differences for at least 1 model, proving that marker style is not cosmetic but consequential. (more details in the Appendix A.)

\paragraph{Effect on model rankings: } Figures~\ref{fig:rank_marker_da2k} and Figure~\ref{fig:rank_marker_spair} show the model accuracy on the default visual marker (red circle, text at the top) along with the delta in performance seen across 16 different visual markers of varying sizes, shapes, colors, and positions. We see that if you change the visual marker, both the absolute accuracy and the relative rankings of models change, which can completely change the overall takeaways. For example, Llama 4 Scout outperforms Gemini 2.5 Flash under the default red marker for relative depth, but under the blue marker condition their ranks reverse, with Gemini 2.5 Flash scoring as the highest performing model. We also see cases of incredibly large differences in accuracy, such as Llama 4 Scout and InternVL for semantic correspondence, where making the text size of the marker smaller results in a 10+\% drop in accuracy and dropping Llama from rank 3 to rank 6. 

\paragraph{Which visual markers matters most?} Figure~\ref{fig:maker_variance_magnitude_da2k} displays the average magnitude of accuracy difference across each marker for the relative depth task, altering text size or label position generally produces larger effects on accuracy, which color and shape of the marker having a less but still noticeable impact. Changing the marker’s color (red to blue) has a large impact on certain models, hinting that those models might be overfit to the color distribution of markers seen in their training or the benchmark (since BLINK’s default is red, some models may have learned to specifically attend to red circles as a prompt cue). Changing the marker’s size and shape or changing the text marker from A/B to 1/2 has a measurable but less prominent effect on accuracy. Notably, there is not a single marker style that was universally best or worst for all models. These idiosyncratic responses point to each model having its own biases in visual prompt processing. Similar plots for semantic correspondence can be found in the Appendix.

\begin{figure*}[h]
    \centering
    \includegraphics[width=\linewidth]{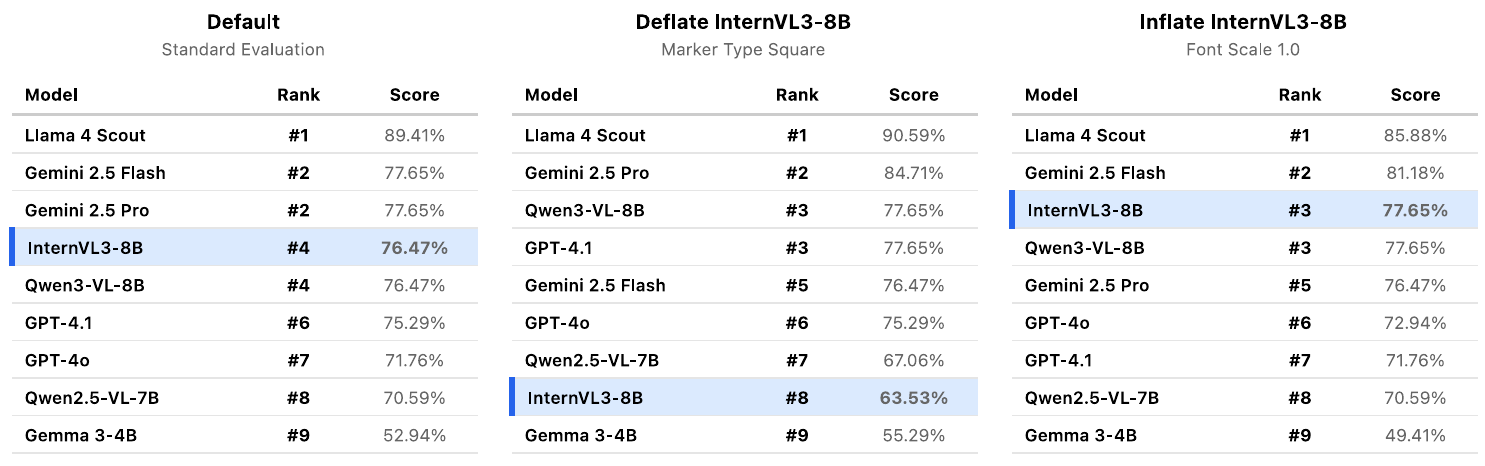}
    \vspace{-1.8em}
    \caption{\textbf{Manipulating leaderboards for InternVL3-8B.} Performance comparison: optimizing for InternVL3-8B's ranking on BLINK relative depth. Leaderboard rank can shift by benign visual prompt choices rather than changes in model capability.}
    \vspace{-0.5em}
    \label{fig:perf_comparison}
\end{figure*}

\paragraph{Manipulating Visually Prompted Leaderboards.}
We explicitly demonstrate that visually prompted leaderboards can be \textit{gamed} by jointly selecting a marker style and an i.i.d.\ split that favor a particular model (Figure~\ref{fig:perf_comparison}). On BLINK relative depth, for example, we lower InternVL3 from rank 4 to rank 8 by changing the marker to a square. Conversely, increasing the marker font size raises its rank to 3, placing it above Gemini 2.5 Pro.  These results show that, without standardized marker conventions, visually prompted evaluations can give misleading impressions of model ability.

\section{Imperceptible differences matter as well}

\begin{figure}
    \centering
    \includegraphics[width=\linewidth]{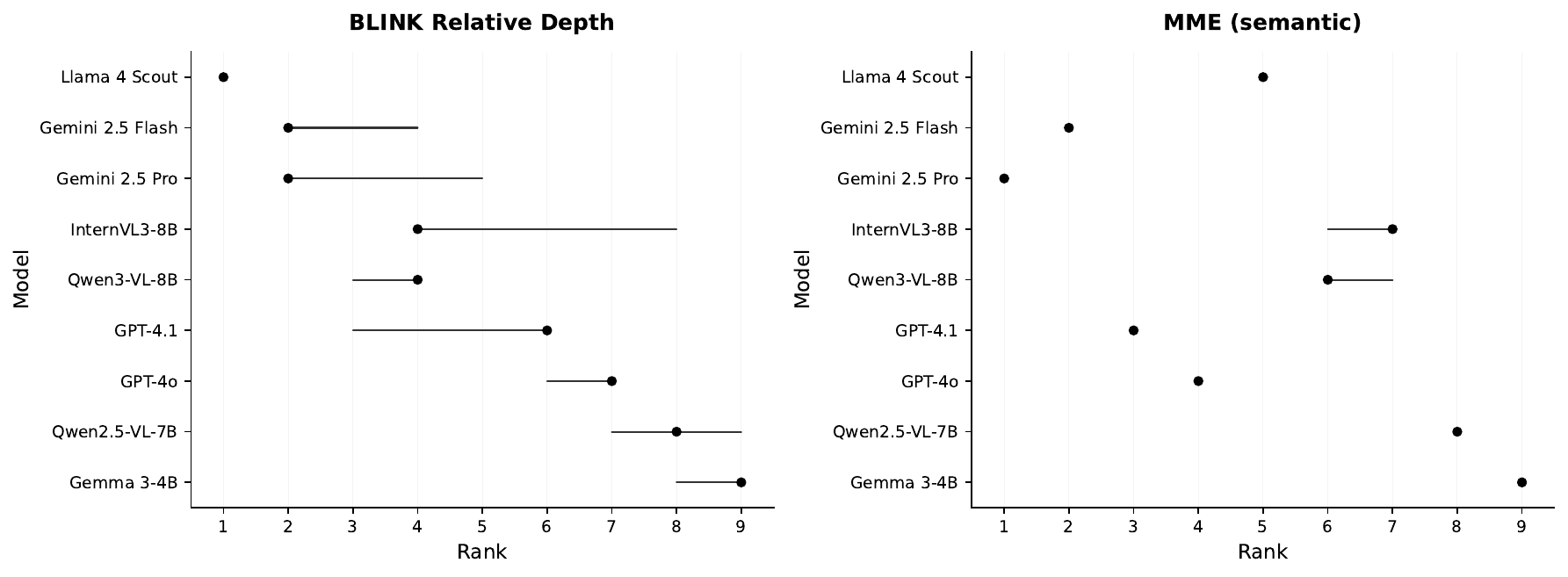}
    \caption{\textbf{Change in rank seen across JPEG compression rates.} Individual dots with no lines indicate models whose rank remains constant across all compression levels. JPEG compression quality substantially reorders model rankings on visually-dependent tasks (BLINK RD), whereas rankings on traditional benchmarks (MME) exhibit greater stability.}
    \vspace{-1.5em}
    \label{fig:jpeg_compression}
\end{figure}

Beyond visually noticeable details such as marker styles, we also extend our investigation to \emph{imperceptible} factors, drawing inspiration from prior work on human perceptual sensitivity and adversarial robustness \cite{zhang2018unreasonable}. In particular, we examine whether common preprocessing operations—such as JPEG compression, which is widely applied in benchmark construction for efficient storage—can subtly affect model performance. Although variations in JPEG quality above a compression level of 70 are largely imperceptible to humans, it remains unclear whether vision–language models (VLMs) exhibit comparable robustness. Moreover, as this issue has not been systematically explored in prior literature, we investigate whether such imperceptible variations affect vision–perception tasks (VPT) differently from conventional knowledge-focused benchmarks.

\textbf{Setup.} We evaluate four different \emph{JPEG compression levels}: default, 70, 80, and 90. Prompts and data splits are kept fixed. To emphasize rank stability rather than absolute performance, Figure \ref{fig:jpeg_compression} reports the change in model ranks for BLINK–Relative Depth and MME. Singular dots mean the model did not change ranks while horizontal lines indicate a change in ranks over different compression rates. The rankings in BLINK RD fluctuate considerably, even for closed-source models such as Gemini~2.5~Pro and GPT-4.1. This sensitivity can likely be attributed to the fact that VPT tasks demand a more fine-grained understanding of visual tokens than semantic-focused tasks; consequently, even subtle pixel-level changes can influence model predictions. In contrast,  MME rankings remain relatively stable across all four compression settings and nine evaluated models, with only a single ranking inversion compared to the pronounced shifts observed in BLINK RD. Given this, we recommend that JPEG quality is be standardized or ideally replaced with lossless formats for VPT benchmarks.

\section{Discussion}

Our results show that visually prompted evaluations introduce non-semantic confounders: marker design, sample choice, and low-level implementation details can all shift accuracies and reorder rankings. This fragility arises both from the benchmarks and from the models themselves.

\noindent To mitigate these issues, we suggest the following:
\begin{itemize}
\item \textbf{Standardize and diversify visual prompts.} Use a clearly specified default marker style, and, whenever possible, report results averaged over a small set of marker variants. For benchmark creators, release clean images together with raw marker coordinates rather than only images with markers baked in, so that alternative prompt designs can be evaluated consistently.
\item \textbf{Enrich test sets from consistent sources.} When feasible, evaluate on larger visually prompted pools constructed from the same underlying data sources and task definitions, rather than relying on a single small split. In our case, we expand BLINK-style relative depth and correspondence tasks with \ours{}-RD and \ours{}-SC using images and annotations from DA2K and SPair-71K, yielding substantially more stable aggregates.
\item \textbf{Adhere to the same realization of low-level settings.} Explicitly standardize and report data-processing and inference choices such as JPEG compression quality, input resolution, and numerical precision (e.g., \texttt{bf16} vs.\ \texttt{fp8}/\texttt{fp16} for self-hosted models). Avoid silent changes across evaluations, and treat models with opaque internals as a separate comparison group. 
\item \textbf{Report uncertainty and rank stability.} Accompany accuracies with confidence intervals and simple rank-stability analyses across markers, splits, or seeds, particularly when the low accuracy tasks have relatively small sample sizes.
When intervals overlap substantially, treat models as tied rather than declaring definitive winners.
\end{itemize}

As a step toward more stable evaluation, we release \ours{}, the scaled-up visually prompted benchmark used in our work, together with multiple marker variants and reference evaluation scripts. With performance of common models on \ours{} shown in~\cref{fig:default_acc}, we show that by reducing variance due to arbitrary design choices, \ours{} makes performance differences more reflective of perceptual ability. While currently limited to depth and correspondence tasks, it provides a foundation for broader analyses and future robustness-aware benchmarks for visual grounding.

\section{Conclusion}

Benchmarks should measure ability, not fragility. Yet our results reveal a gap: visually prompted evaluations ``fall short of this standard'': change a marker’s color, shift its label, compress an image differently, and entire leaderboards reshuffle. These shifts are not noise but structural weaknesses, revealing that today's perception-focused VLM benchmarks are far more sensitive than the field assumes. Although our demonstrations center on BLINK-style tasks, the pattern is unlikely to be isolated. Any benchmark that depends on explicit visual markup or fine-grained spatial cues risks similar instability. If leaderboards can be flipped by choices orthogonal to task semantics, they cannot be trusted to track genuine progress. To address this, we recommend evaluations should diversify visual prompts, report variance in addition to scores, and standardize low-level settings that silently influence results. \ours{} is a step in that direction, offering larger, marker-diverse test sets that reduce incidental variance. Stable measurement is a prerequisite for meaningful comparison; until then, visually prompted leaderboards may be telling us more about their construction than about the models they rank.

\noindent \textbf{Acknowledgments.} We thank David Chan, Jacob Steinhardt, and Joey Gonzalez for helpful discussions and feedback on statistical tests. We also thank Junyi Zhang, Melanie Sclar, and Grace Luo for suggestions on experimental design. Lastly, we thank Jitendra Malik, Jiaxin Ge, and Anne Harrington for thoughtful feedback on the manuscript.

\bibliography{iclr2026_conference}
\bibliographystyle{iclr2026_conference}

\clearpage
\appendix
\vspace*{-\topskip}   %
\begin{center}
  {\LARGE\bfseries Supplementary Materials}
\end{center}
\vspace{1em}
\section{Marker style significance}

\begin{figure*}[h]
    \centering
    \begin{subfigure}{0.48\linewidth}
        \centering
        \includegraphics[width=\linewidth]{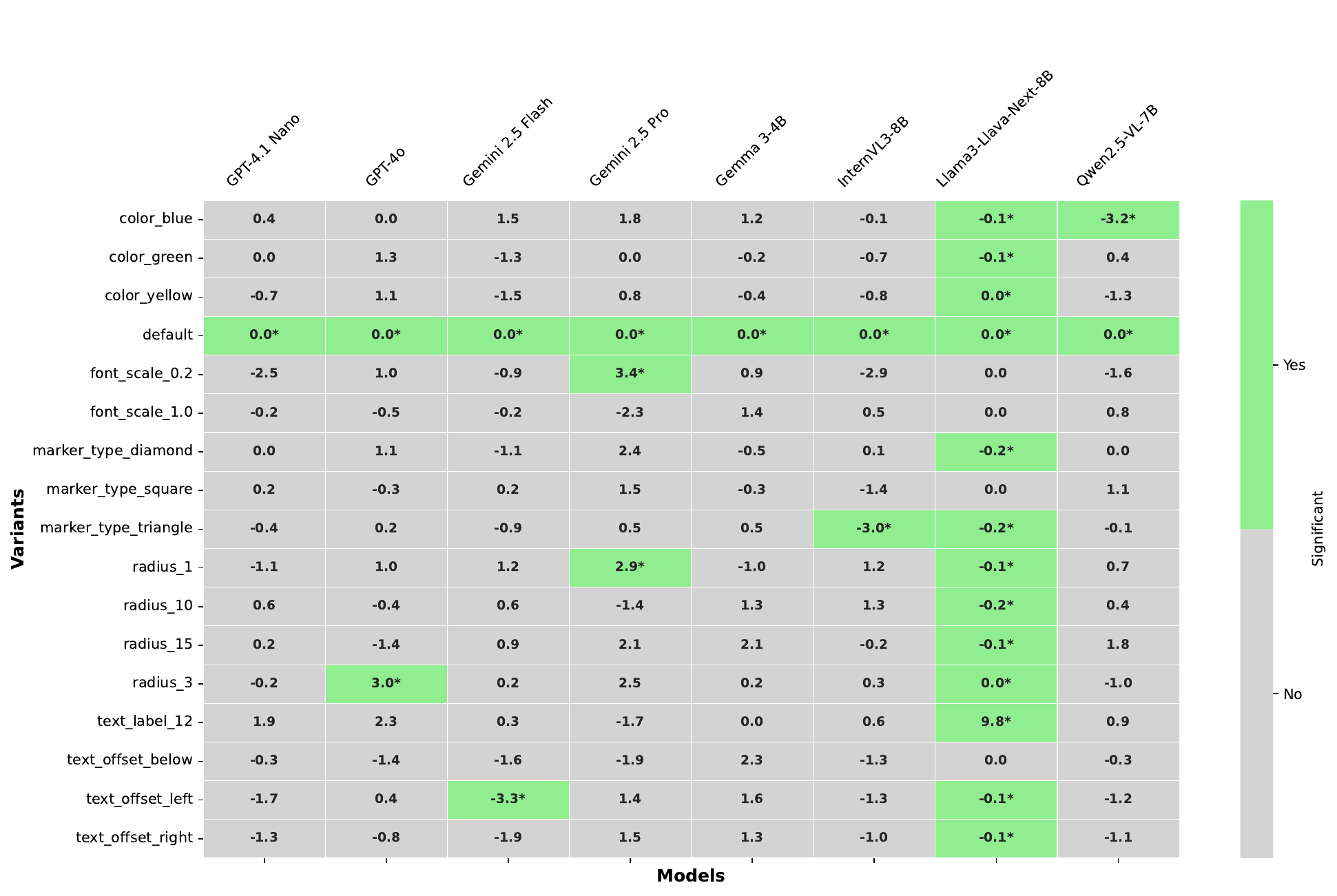}
        \caption{VPBench-Relative Depth}
        \label{fig:significance_da2k}
    \end{subfigure}
    \hfill
    \begin{subfigure}{0.48\linewidth}
        \centering
        \includegraphics[width=\linewidth]{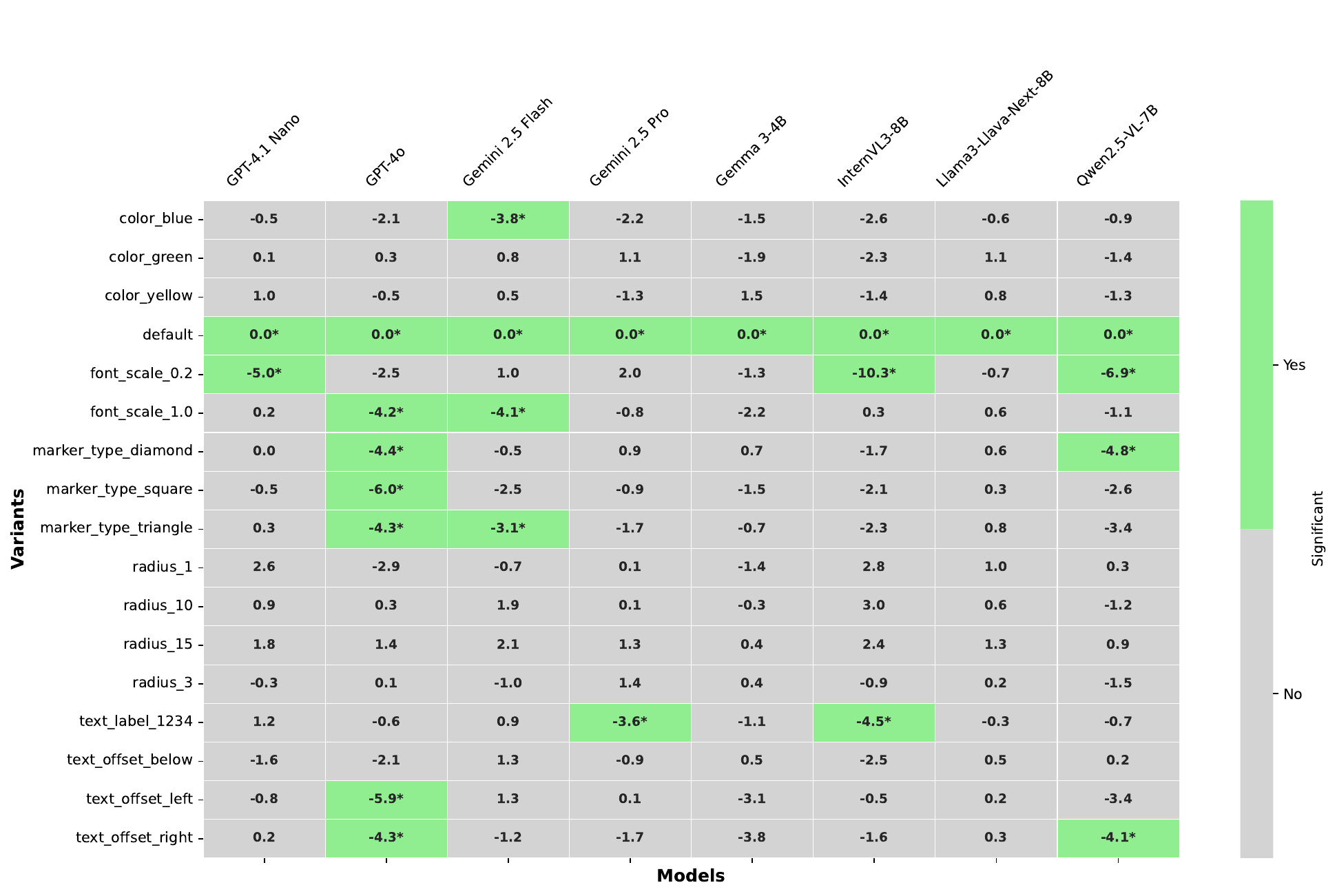}
        \caption{VPBench-Semantic Correspondence}
        \label{fig:significance_spair}
    \end{subfigure}

    \caption{\textbf{Significance plots of marker variants.} Green indicates statistical significance under paired bootstrap. We see that each model has at least 1 marker variant which produces a statistically significant difference in accuracy compared to the default marker.}
    \label{fig:significance_cis}
\end{figure*}

\begin{figure}
    \centering
    \includegraphics[width=\linewidth,trim={1cm 0cm 4cm 1cm},clip]{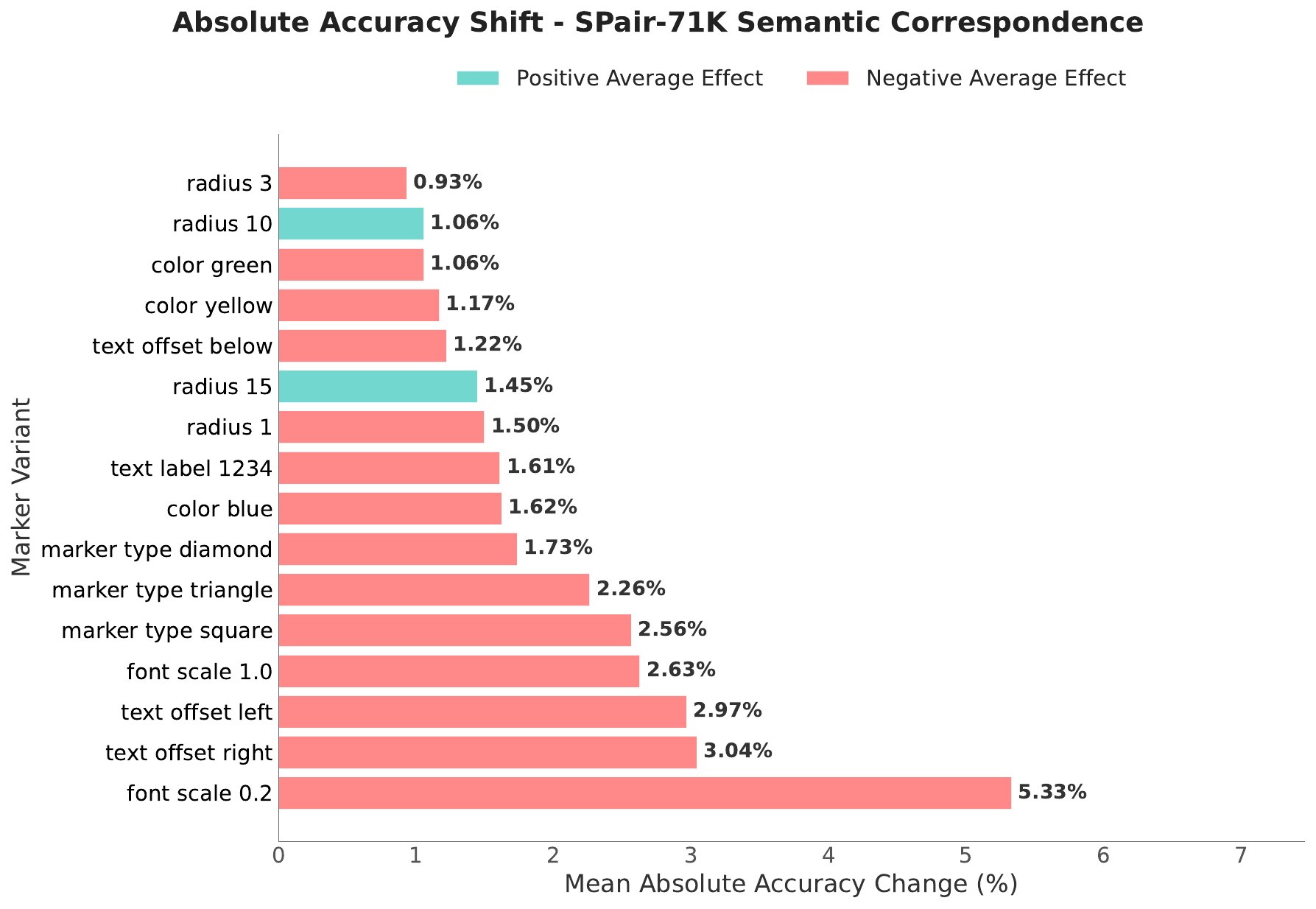}

    \caption{\textbf{Absolute marker impact.} Mean absolute accuracy shift on VPBench-Semantic Correspondence induced by each marker variant. Variants which alter the text component of the visual marker typically result in the largest accuracy shifts. }
    \label{fig:maker_variance_magnitude_spair}
\end{figure}

\begin{figure*}
    \centering
    \includegraphics[width=\linewidth, trim=1cm 0.5cm 2cm 1cm, clip]{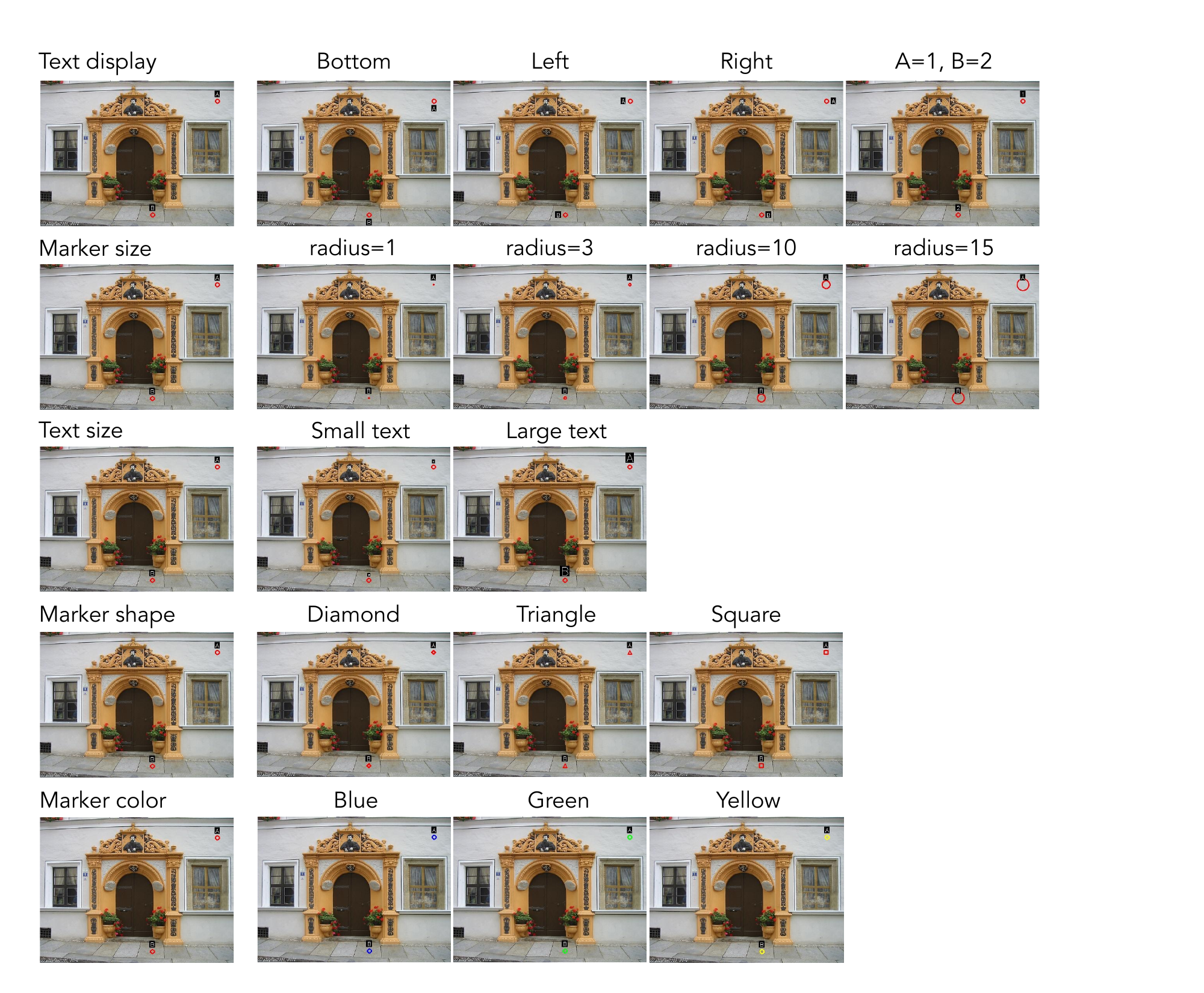}
    \caption{\textbf{Visual Marker Variants.} We explore 16 different visual markers in Section~\ref{sec:markers} of the main paper. }
    \label{fig:marker_choice_example}
\end{figure*}

Figure~\ref{fig:significance_cis} reports which marker-induced differences on \ours{}-RD and \ours{}-SC are statistically significant under paired confidence intervals, we can see that all the models have at least one style that is statistically significant, justify the independence of the marker style variance over the data variance. We additionally show the mean accuracy shift per marker on \ours{}-SC in Figure~\ref{fig:maker_variance_magnitude_spair}, in which we see that similar to the results from \ours{}-RD in the main paper (Figure~\ref{fig:maker_variance_magnitude_da2k}), the marker variants with the largest effects are the ones which involve changing the placement, size, and representation of the text of the marker. 
In Figure~\ref{fig:marker_choice_example}, we illustrate the marker variants used in our experiments.
\begin{figure*}[t]
    \centering

    \begin{subfigure}{0.4\linewidth}
        \centering
        \includegraphics[width=0.75\linewidth, trim=0 0pt 0pt 23pt, clip]{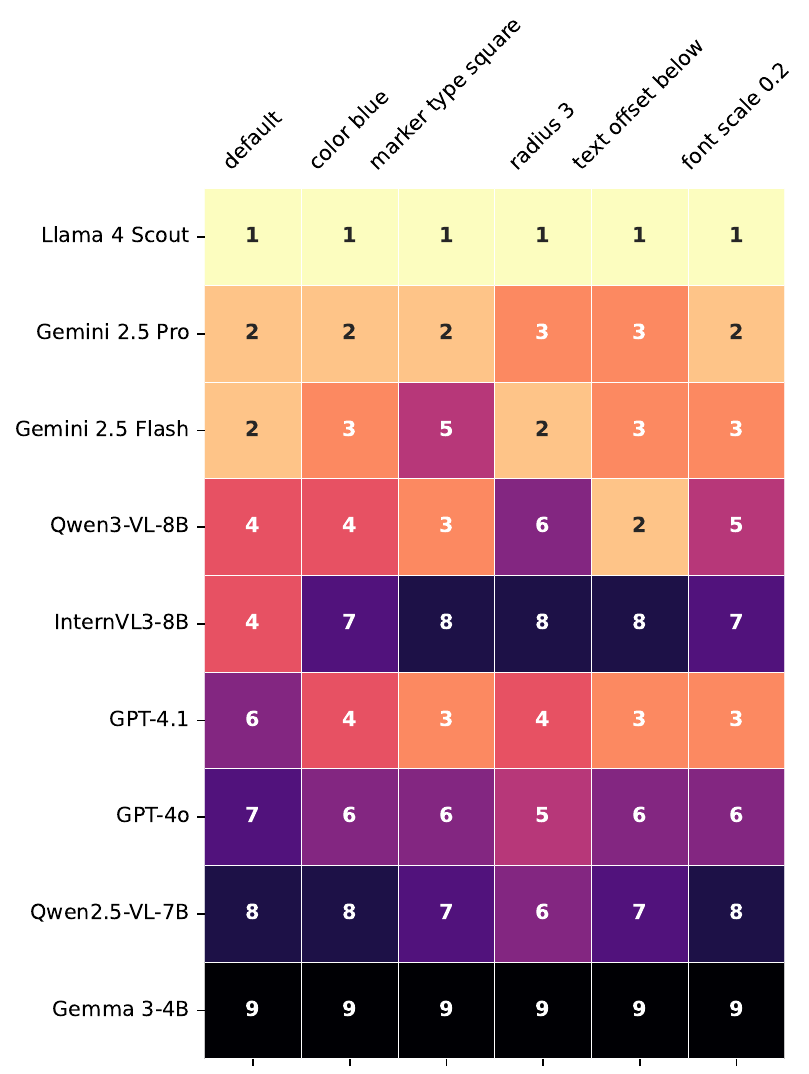}
        \caption{BLINK RD}
        \label{fig:blink_rd_marker_rank}
    \end{subfigure}
    \hfill
    \begin{subfigure}{0.4\linewidth}
        \centering
        \includegraphics[width=0.75\linewidth, trim=0 0pt 0pt 23pt, clip]{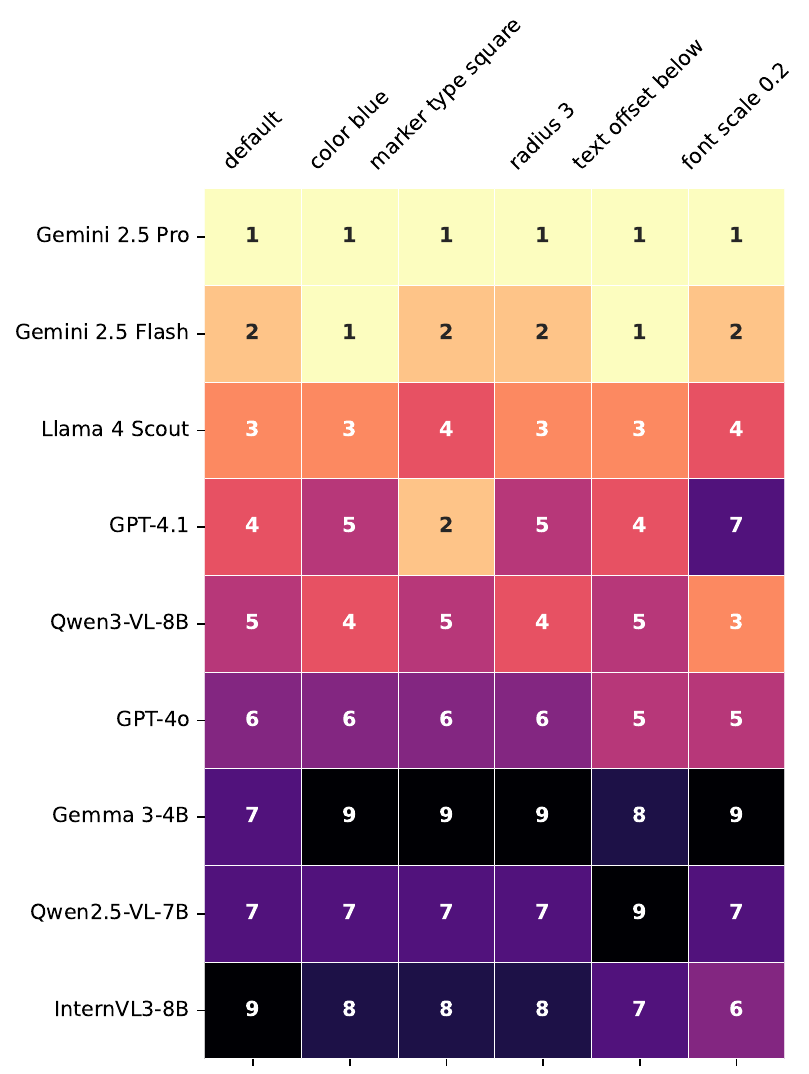}
        \caption{BLINK SC}
        \label{fig:blink_sc_marker_rank}
    \end{subfigure}

    \vspace{0.5em}

    \begin{subfigure}{0.4\linewidth}
        \centering
        \includegraphics[width=0.75\linewidth, trim=0 0pt 0pt 23pt, clip]{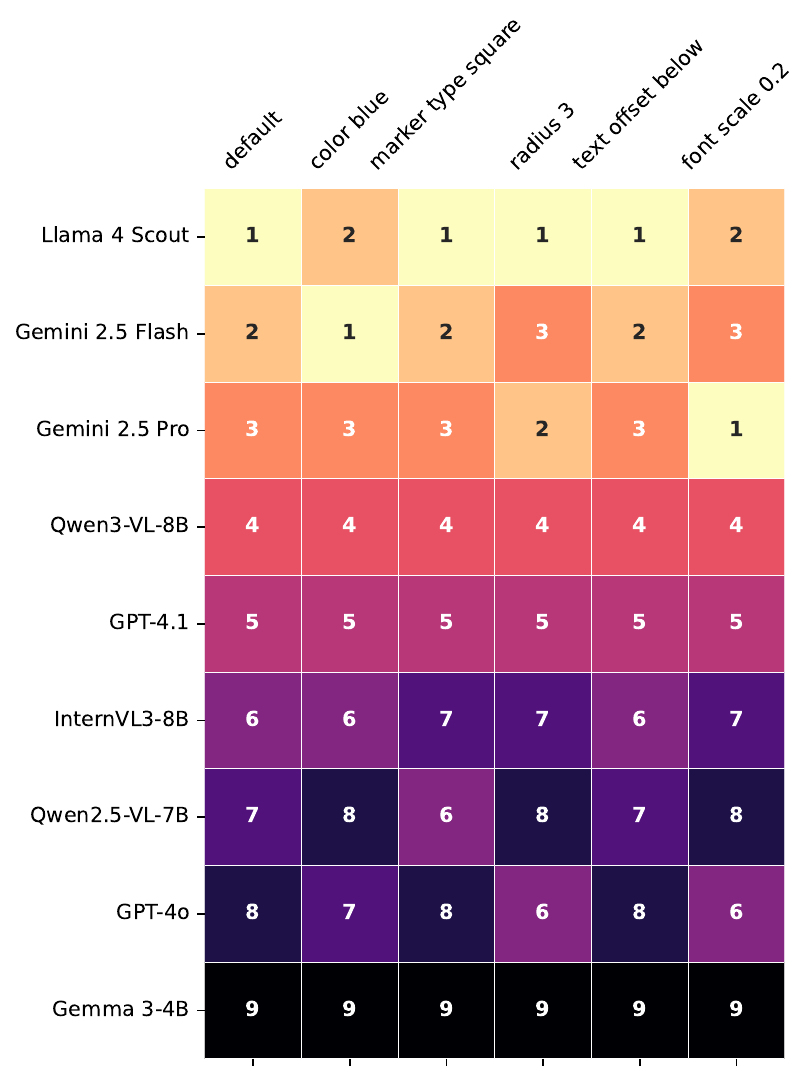}
        \caption{\ours{} RD}
        \label{fig:da2k_marker_rank}
    \end{subfigure}
    \hfill
    \begin{subfigure}{0.4\linewidth}
        \centering
        \includegraphics[width=0.75\linewidth, trim=0 0pt 0pt 23pt, clip]{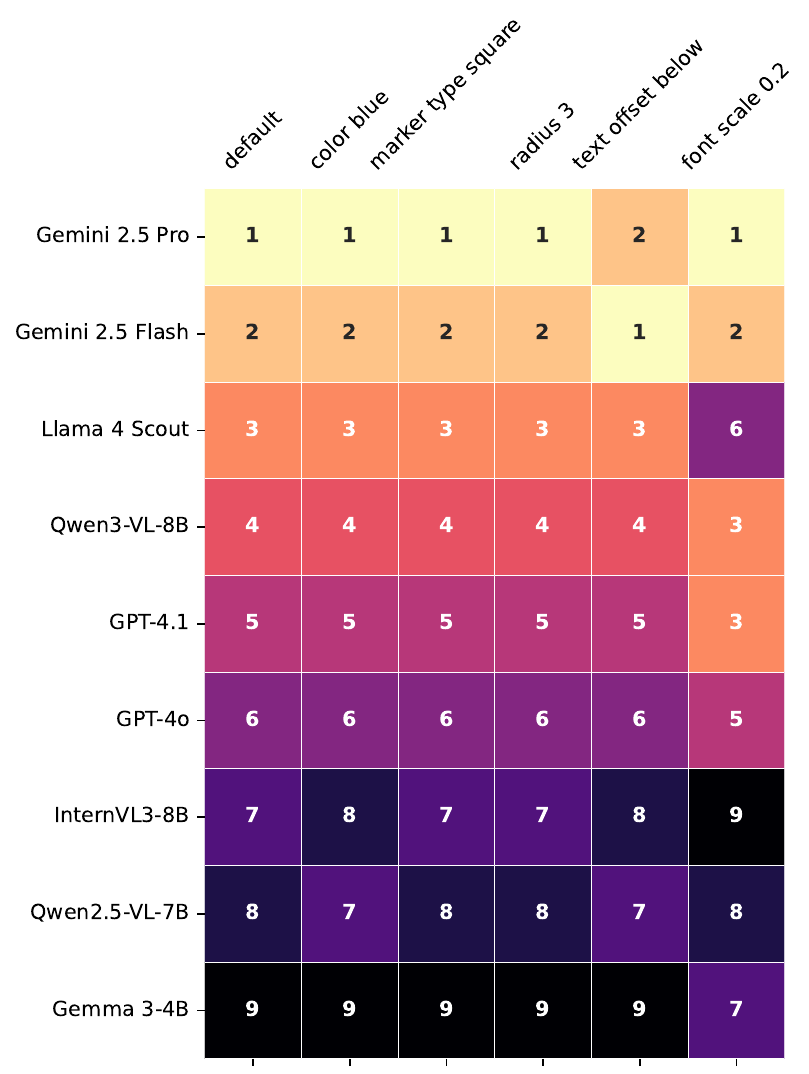}
        \caption{\ours{} SC}
        \label{fig:spair_marker_rank}
    \end{subfigure}

    \caption{\textbf{Change in rank for different marker styles} across BLINK RD, BLINK SC, \ours{} RD, and \ours{} SC. For each dataset we see large fluctuations in rank across marker types, indicating that these tasks are highly sensitive to small visual changes.}
    \label{fig:marker_style_rank_all}
\end{figure*}

\section{Change in accuracies on all marker styles}
\begin{figure*}
    \centering
    \includegraphics[width=0.8\linewidth, trim=0cm 0cm 4cm 2cm, clip]{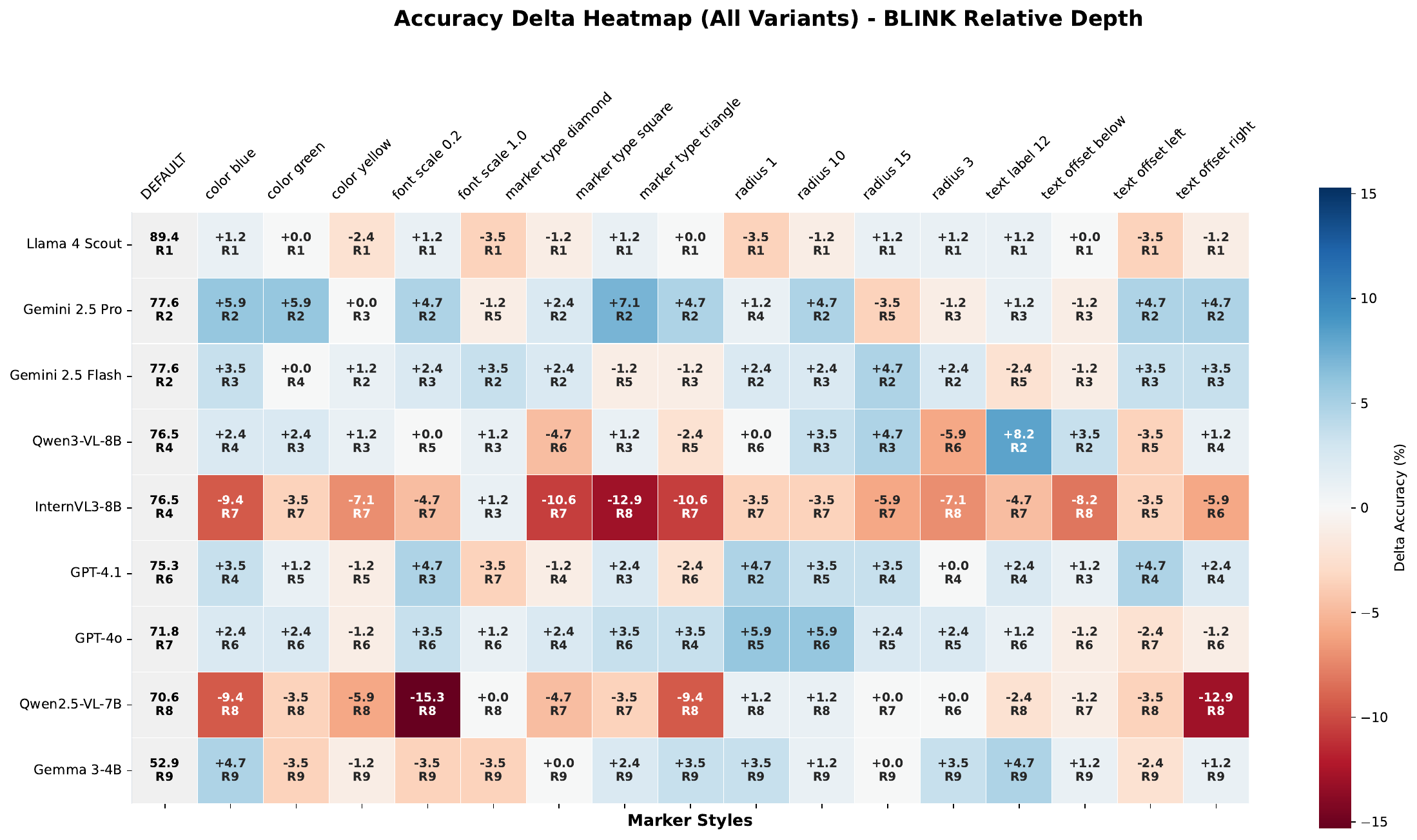}
    \caption{Change in accuracy and rank for different marker styles in BLINK relative depth task.}
    \label{fig:blink_rd_marker}
\end{figure*}

\begin{figure*}
    \centering
    \includegraphics[width=0.8\linewidth, trim=0cm 0cm 4cm 2cm, clip]{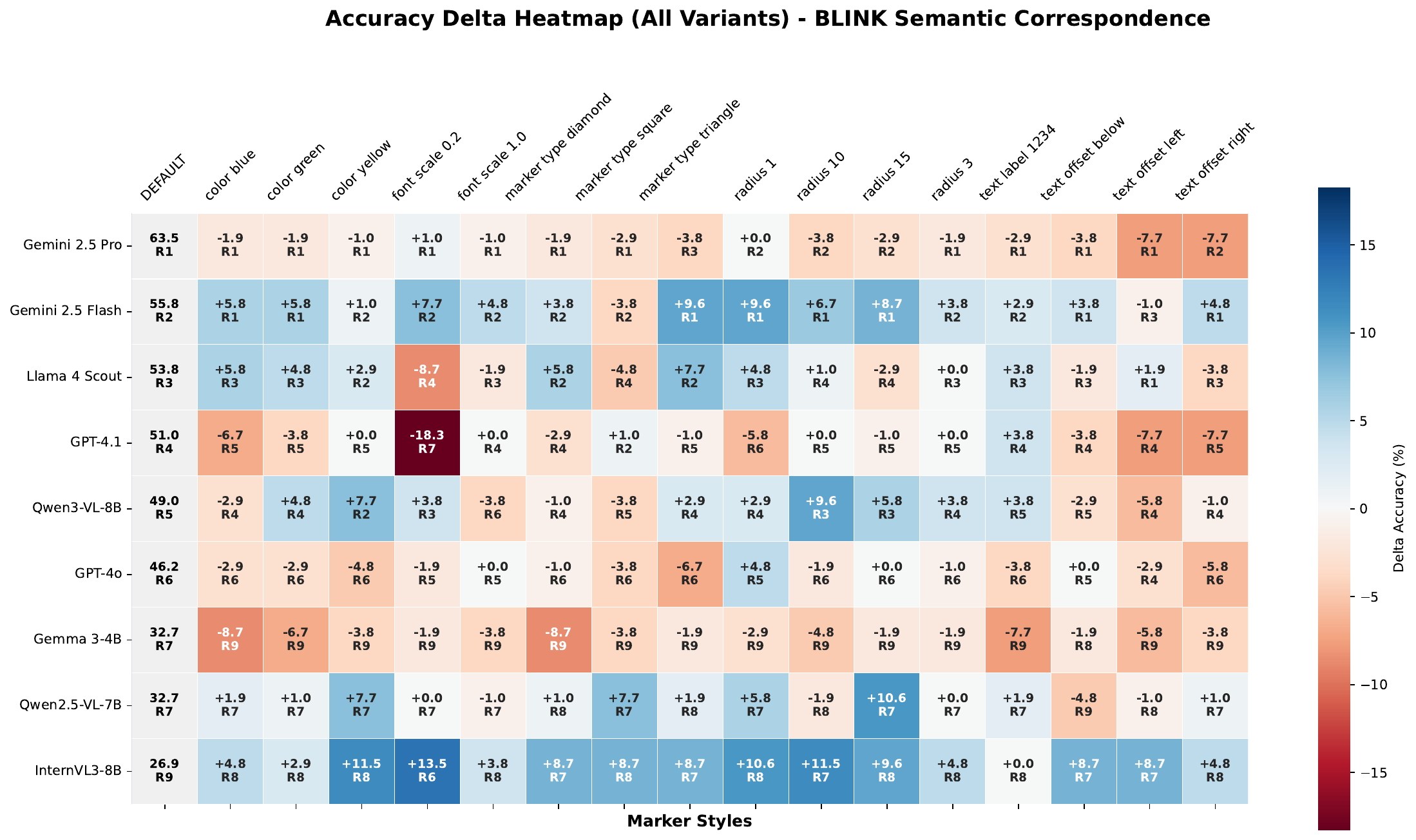}
    \caption{Change in accuracy and rank for different marker styles in BLINK semantic correspondence task.}
    \label{fig:blink_sc_marker}
\end{figure*}

\begin{figure*}
    \centering
    \includegraphics[width=0.8\linewidth, trim=0cm 0cm 4cm 2cm, clip]{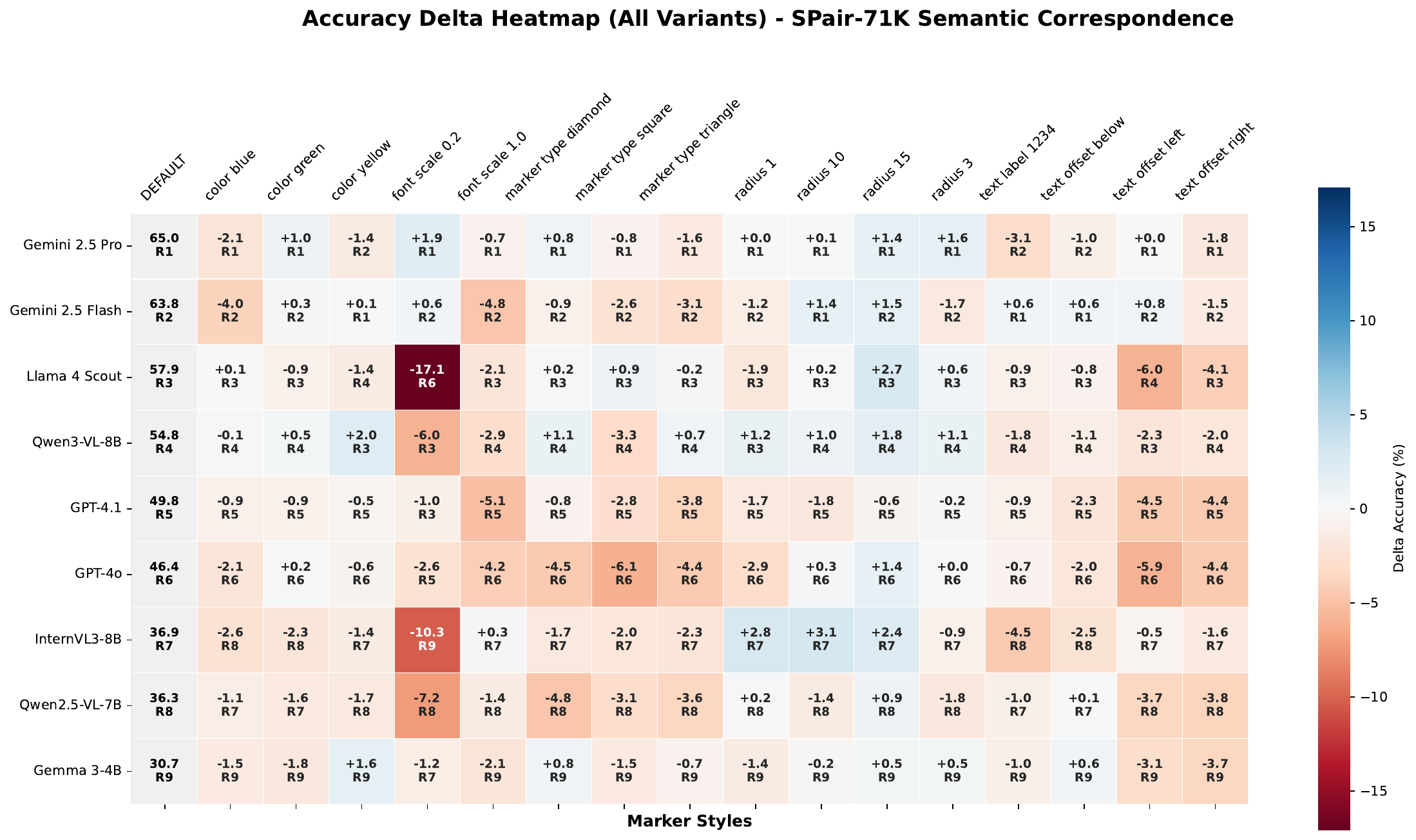}
    \caption{Change in accuracy and rank for different marker styles in VPBench-Semantic Correspondence.}
    \label{fig:spair_marker}
\end{figure*}

\begin{figure*}
    \centering
    \includegraphics[width=0.8\linewidth, trim=0cm 0cm 4cm 2cm, clip]{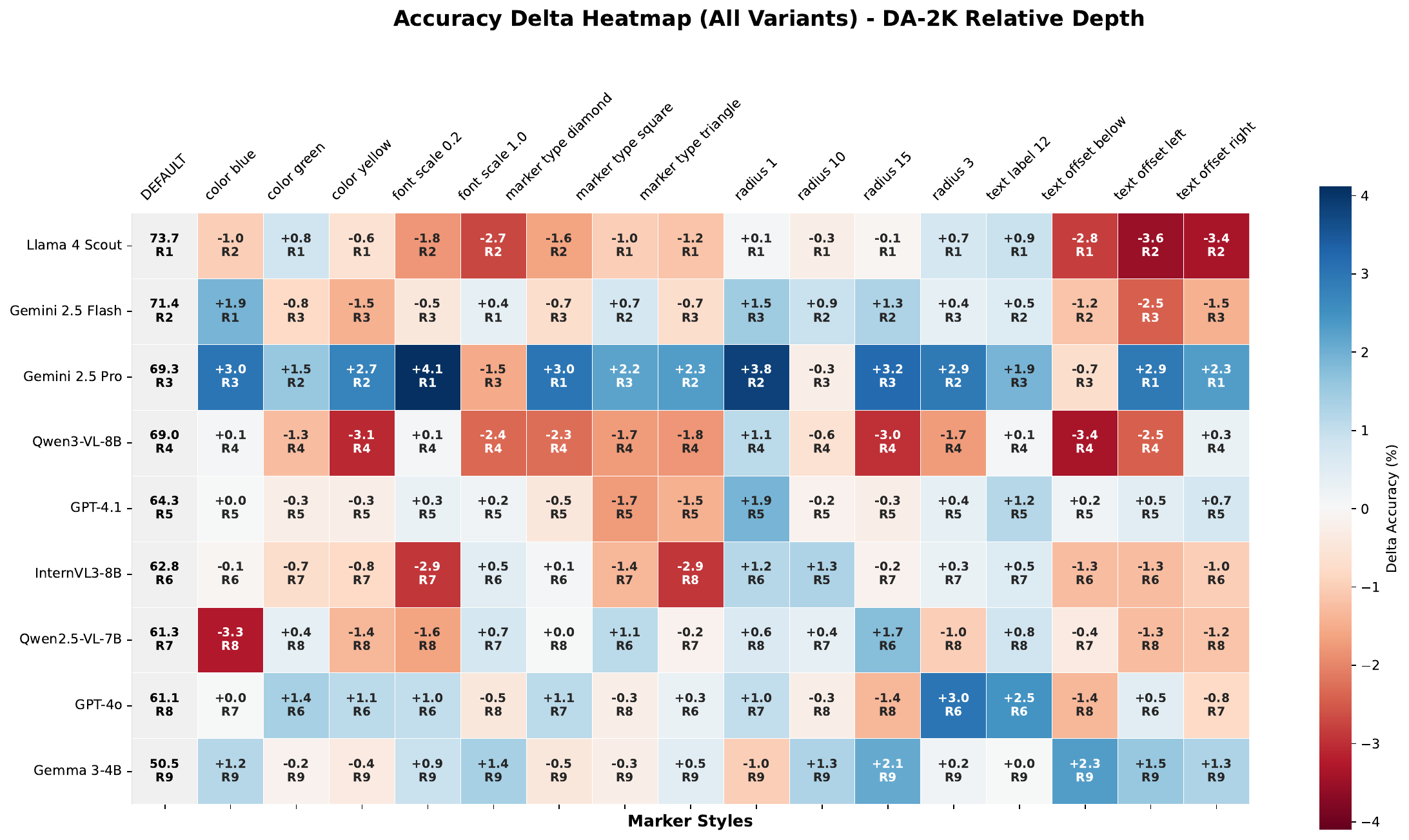}
    \caption{Change in accuracy and rank for different marker styles in VPBench-Relative Depth.}
    \label{fig:da2k_marker}
\end{figure*}

Below we show the full change in accuracy and rank for all marker styles across BLINK Relative Depth (\cref{fig:blink_rd_marker}), BLINK Semantic Correspondence (\cref{fig:blink_sc_marker}), \ours{}-SC (\cref{fig:spair_marker}) and \ours{}-RD (\cref{fig:da2k_marker}), with the induced changes in model ranking isolated in Figures~\ref{fig:blink_rd_marker_rank}-\ref{fig:spair_marker_rank}.

\paragraph{BLINK Relative Depth and Semantic Correspondence.} For BLINK relative depth, changing only the marker style (color, size, shape, or label layout) yields drastic accuracy shifts, sometimes up to roughly 15\% for individual models, as shown in the accuracy heatmap in \cref{fig:blink_rd_marker}. These shifts are large enough to reorder nearby models in the leaderboard, with several mid-ranked systems moving up or down multiple positions across marker variants (\cref{fig:blink_rd_marker_rank}). BLINK semantic correspondence shows a similar pattern: accuracy often changes by more than 10\% under different marker styles (\cref{fig:blink_sc_marker}), and these shifts again reorder models with similar default performance (\cref{fig:blink_sc_marker_rank}), so marker design alone can change which model appears to perform best on both BLINK tasks.

\paragraph{\ours{} Semantic Correspondence.}
For \ours{}-SC, changing the marker style shifts model accuracies in systematic ways, with some variants consistently helping or hurting broad groups of models, as shown in \cref{fig:spair_marker}. These shifts are also large enough to change the relative ordering of mid-ranked systems, with multiple models swapping positions across marker styles in \cref{fig:spair_marker_rank}.

\paragraph{\ours{} Relative Depth.}
On \ours{}-RD, marker style changes also lead to clear accuracy shifts for most models, though typically smaller than on BLINK potentially due to its larger data size, as shown in \cref{fig:da2k_marker}. Variants that alter marker size or label layout tend to have the strongest effect, while pure color or shape changes are milder but still noticeable. These differences are often sufficient to reorder models that are close in performance, especially away from the very top of the leaderboard, as illustrated in the rank changes in \cref{fig:da2k_marker_rank} and the significance analysis in \cref{fig:significance_cis}.

\end{document}